\documentclass[letterpaper]{article} 
\usepackage{aaai24}  
\usepackage{times}  
\usepackage{helvet}  
\usepackage{courier}  
\usepackage[hyphens]{url}  
\usepackage{graphicx} 
\usepackage{natbib}  
\usepackage{caption} 
\usepackage{algorithm}

\usepackage{algorithm}

\usepackage{amsmath}
\usepackage{booktabs}       
\usepackage{amsfonts}       
\usepackage{xcolor}         
\usepackage{amsthm}
\newtheorem{definition}{Definition}

\usepackage{graphicx}
\newtheorem{prop}{Proposition}
\usepackage{algorithm}
\usepackage{algpseudocode}
\usepackage{multirow}
\usepackage{ctable} 
\usepackage{mathtools}
\usepackage{subcaption}
\usepackage{verbatim}
\usepackage{lipsum,booktabs}

\DeclareMathOperator{\E}{\mathbb{E}}

\usepackage{newfloat}
\usepackage{listings}
\DeclareCaptionStyle{ruled}{labelfont=normalfont,labelsep=colon,strut=off} 
\floatstyle{ruled}
\newfloat{listing}{tb}{lst}{}
\floatname{listing}{Listing}
\title{Make Prompts Adaptable: Bayesian Modeling \\ for Vision-Language Prompt Learning with Data-Dependent Prior}
\author {
    Youngjae Cho\textsuperscript{\rm 1},
   HeeSun Bae\textsuperscript{\rm 1},
    Seungjae Shin \textsuperscript{\rm 1},Yeo Dong Youn \textsuperscript{\rm 2}, Weonyoung Joo \textsuperscript{\rm 3} \\ Il-Chul Moon \textsuperscript{\rm 1}
}
\affiliations {
    \textsuperscript{\rm 1} KAIST\\
    \textsuperscript{\rm 2} Seoul National University\\
    \textsuperscript{\rm 3} Ewha Womans University

    \{genius.cho, cat2507, tmdwo0910\}@kaist.ac.kr, dbsduehd95@snu.ac.kr, weonyoungjoo@ewha.ac.kr, icmoon@kaist.ac.kr
}

\begin{document}

\maketitle

\begin{abstract}
    Recent Vision-Language Pretrained (VLP) models have become the backbone for many downstream tasks, but they are utilized as frozen model without learning. Prompt learning is a method to improve the pre-trained VLP model by adding a learnable context vector to the inputs of the text encoder. In a few-shot learning scenario of the downstream task, MLE training can lead the context vector to over-fit dominant image features in the training data. This overfitting can potentially harm the generalization ability, especially in the presence of a distribution shift between the training and test dataset. This paper presents a Bayesian-based framework of prompt learning, which could alleviate the over-fitting issues on few-shot learning application and increase the adaptability of prompts on unseen instances. Specifically, modeling data-dependent prior enhances the adaptability of text features for both seen and unseen image features without the trade-off of performance between them. Based on the Bayesian framework, we utilize the Wasserstein Gradient Flow in the estimation of our target posterior distribution, which enables our prompt to be flexible in capturing the complex modes of image features. We demonstrate the effectiveness of our method on benchmark datasets for several experiments by showing statistically significant improvements on performance compared to existing methods. The code is available at \url{https://github.com/youngjae-cho/APP}.

\end{abstract}

\section{Introduction}
Recently, Vision-Language Pretrained models (VLP) \cite{radford2021learning,jia2021scaling} have been used as backbones for various downstream tasks \cite{shen2022how,ruan2022optimal}, and the pre-trained models have shown successful adaptation. 
Since these pre-trained models are used as-is in downstream tasks, \textit{prompt learning} adds a context vector to the input of pre-trained model, so the context vector becomes the learnable parameter to improve the representation from the pre-trained model \cite{zhou2022learning} for the downstream task. For instance, a text input is concatenated to a context vector, and the new text input could be fed to the text encoder. The learning of context vector comes from the back-propagation after the feed-forward of the concatenated text input. Since there is only a single context vector without being conditioned by inputs, the inferred value of context vector becomes a static single context defined for the given downstream task. 

Whereas improving parameter-frozen VLP models by additional input context vector is a feasible solution, it can potentially overfit to a dense area of image features in few-shot learning. Since text features are hard to capture the multi-modes of image features in MLE training, it could fail to infer the minor area of image features, eventually degrading the performance. In addition, MLE training undermines the generalization capability of VLP models especially when there is a distribution shift between the training and testing \cite{zhou2022conditional}. Although several input-conditioned prompt learning \cite{zhou2022conditional,derakhshani2023bayesian} tried to generalize unseen datasets, it inevitably undermines the performance of seen datasets.


\begin{figure*}
    \centering
    \begin{subfigure}{0.525\textwidth}
        \includegraphics[width=\textwidth]{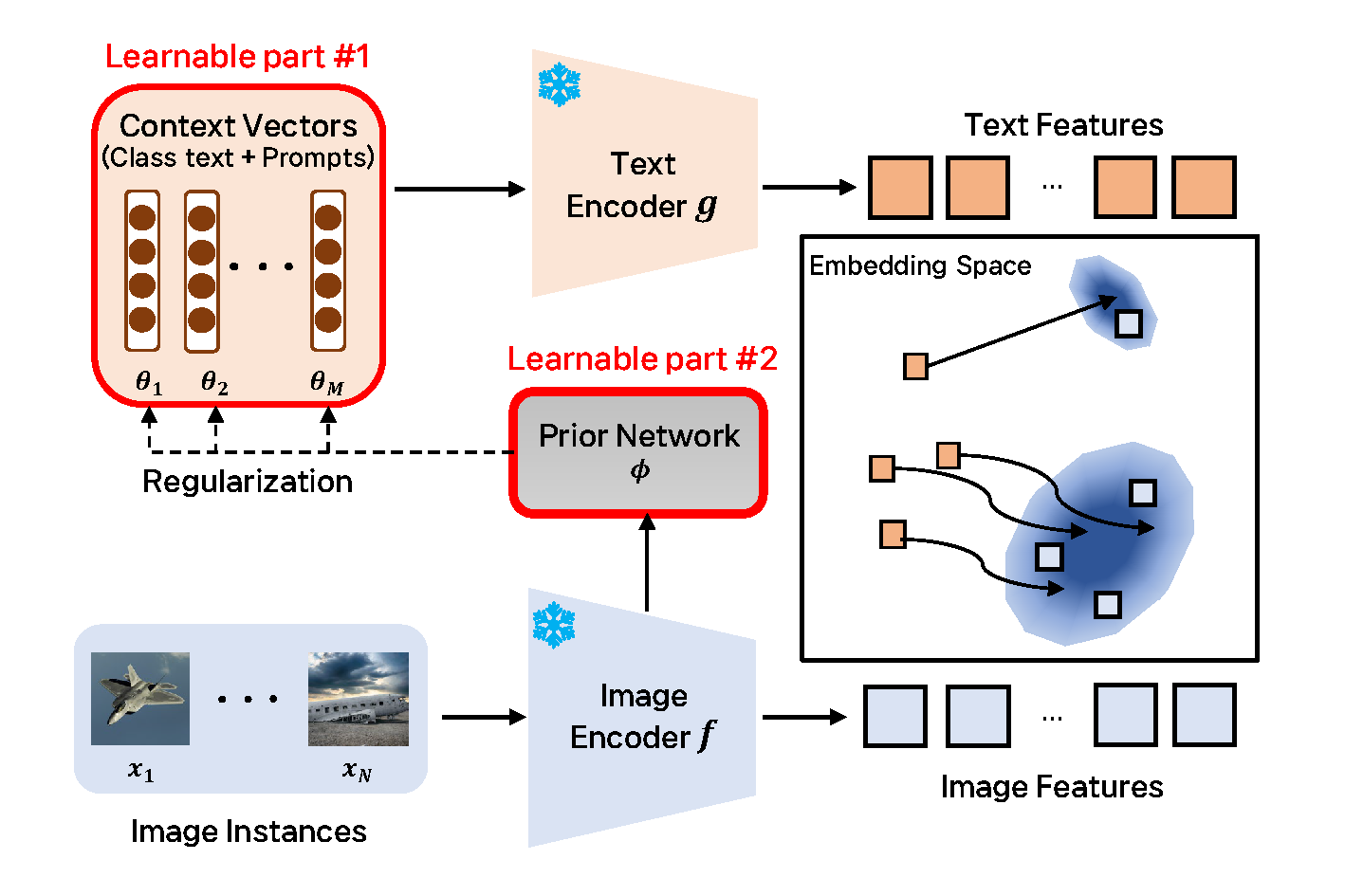}
        \caption{Structure view of APP}
        \label{fig:subfig1}
    \end{subfigure}
    \hfill 
    \begin{subfigure}{0.425\textwidth}
\includegraphics[width=\textwidth]{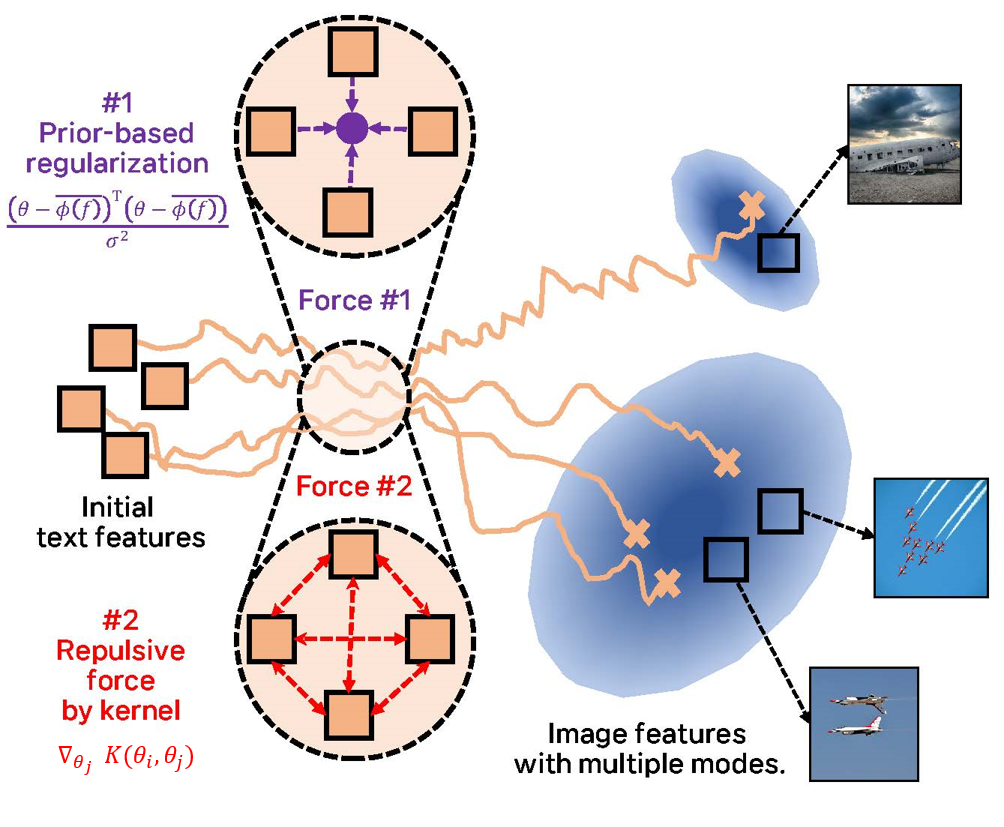}
        \caption{Learning dynamics of APP}
        \label{fig:subfig2}
    \end{subfigure}
    \caption{Structure (left) and learning dynamics (right) of APP. Multiple context vectors are particles of approximated distribution and image conditioned prior can guide the context vector to capture the multi modes.}
    \vskip-0.1in
    \label{fig:whole}
\end{figure*}

To alleviate the impact of uncertainty arising from a few-shot learning scenario, our paper proposes Adaptive Particle-based Prompt Learning (APP), which utilizes a Bayesian inference for prompt learning with a data-dependent prior as shown in Figure \ref{fig:whole}. Through regularization using this data-dependent prior, the context vector is directed toward capturing the diverse modes in image features among the seen data instances. Additionally, we approximate the posterior distribution via Wasserstein Gradient Flow to enhance the flexibility of our text features to infer the complex image features. Furthermore, we extend the modeling of the data-dependent prior to unseen test instances to adapt the distribution shift. This adaptation of the context vector to the unseen instances enhances the model's resilience in the face of distribution shifts, providing robustness to these variations.

We summarize our contributions in two aspects.
\begin{enumerate}
\item Enhancing Flexibility of Prompt: By approximating prompt posterior with Wasserstein Gradient Flow, our context vectors can be more flexibly utilized to infer the complex image feature spaces.
\item Enhancing Adaptability of Prompt: By modeling data-dependent prior based on the image feature information, text features capture the multi-modes of seen image features and adapt to unseen image features, which leads to the improved performances of seen and unseen datasets without trade-off. 

\end{enumerate}


\section{Preliminary}
\subsection{Formulation of Prompt Learning}
\paragraph{Deterministic Prompt Learning}
The goal of prompt learning, i.e. CoOp \cite{zhou2022learning}, is to facilitate adaptation of a given Vision-Language Pretrained model for a target task through learning arbitrary context vectors. In this setting, the pre-trained VLP model is frozen under the prompt learning task, and the additional learnable input is added to text inputs. Finally, the learning gradient is obtained from adapting the target task, a.k.a. downstream task. When we define $(X,Y)$ as a pair of image and its label (i.e. text phrase), Eq. \ref{eq:1} shows a log-likelihood of prompt learning in the image classification.
\begin{align}
\label{eq:1}
  &\mathcal{L}_{CE}(\theta,X,Y)=-\log p (Y|X,\theta) \notag \\  &= -\sum_{i=1}^{N} \{ \log( {\exp(\text{sim}(g(\theta,y_i),f(x_i))/ \tau) \over\sum_{k=1}^{C}\exp(\text{sim}(g(\theta,y_k),f(x_i))/ \tau) })   \}
\end{align}
Here, $f$ and $g$ are image and text encoders from VLP models, respectively; and they are frozen in prompt learning. Therefore, the only learnable part is $\theta$, which is a context vector. $\theta \in \mathbb{R}^d$ is learned as a unique vector for each downstream task without discriminating the data instances. Since $g$ is often implemented as a transformer to take sequential inputs of any length, $g$ does not need to be modified to accept $y$ and $\theta$. Additionally, $\text{sim}(\cdot)$ represents cosine similarity, and $\tau$ is the annealing temperature.

From Eq. \ref{eq:1}, we define the prompt as $\{ \theta,y_i \}$, which is the concatenation of the context vector and the label. By minimizing Eq. \ref{eq:1}, $\theta$ is learned to maximize the alignment of the space between the image feature $f(x)$ and the text feature $g( \theta,y_i)$. Therefore, the learnable part of prompt learning contributes from the input space side, rather than the frozen VLP model parameters. There were some follow-up researches on CoOp. For example, CoCoOp \cite{zhou2022conditional} extended CoOp by learning the image-conditional context vector as $\theta+\phi(f(x_i))$, where $\phi$ is a neural network to map image feature to the prompt space.   



\paragraph{Probabilistic Prompt Learning}
Given a few instance of training dataset, the point estimate of text feature given prompt is hard to capture unseen image feature. ProDA \cite{lu2022prompt} is the first probabilistic model, where the text feature-given prompt is approximated as a Gaussian distribution with a regularizer to enhance the diversity of text feature. PLOT \cite{chen2023plot} formulates the prompt learning as optimal transport, where image and text features are defined as a discrete distribution by Dirac measure. The text features, given as multiple prompts, are assigned to the locality of image features to learn diverse semantics.



\paragraph{Bayesian Probabilistic Prompt Learning}
Since MLE training can induce overfitting with a few training dataset, Bayesian inference is needed to mitigate the high data variance from such a limited dataset.
BPL \cite{derakhshani2023bayesian} is the first prompt learning model from the view of Bayesian inference, which uses variational inference to approximate the posterior distribution with a parameterized Gaussian distribution. The objective function of BPL is defined as follows:
\begin{equation}
\label{eq:2}
    \E_{q(r|X)} [ \log (p(Y|X,\theta,r)) ]-D_{KL}(q(r|X)||p(r))
\end{equation}
where $r$ is a random variable conditioned on $X$, and $r$ is added to a deterministic $\theta$ to turn it into a random variable, which is a reparameterization trick. In detail, the distribution of $r$ is given by $q(r|X)$, which is a Gaussian distribution parameterized by $m(f(X))$ and $\Sigma(f(X))$, where $m(f(X))$ and $\Sigma(f(X))$ are functions of the image feature $f(X)$. The prior distribution over $r$ is $p(r)$, which is a standard Gaussian distribution $N(0,I)$.


\subsection{Wasserstein Gradient Flow}
Bayesian inference is a solution to mitigate the uncertainty from modeling the posterior distribution of parameters. Often, the hurdle of the Bayesian inference is the inference of the posterior distribution, which could be difficult, i.e. modeling either prior or likelihood to be flexible without being conjugate to each other. For improving the posterior distribution to be flexible and multi-modal, we need an inference tool for this complex posterior distribution.

For instance, the JKO scheme \cite{doi:10.1137/S0036141096303359} interprets variational inference as gradient flow, which minimizes the KL divergence between the variational distribution $q$ and the true posterior distribution $\pi \propto \exp(-V(\theta))$  in Wasserstein Space, where $V(\theta)$ is an energy function of posterior distribution. The learning objective of this variational posterior inference $F(q)$ becomes the KL Divergence as follows.
\begin{equation}
    \label{eq:3}
    F(q) \coloneqq D_{KL}(q||\pi) \approx \E_{q} [V(\theta) + \log q]
\end{equation}
To compute the steepest gradient of $F(q)$, we define the Wasserstein Gradient Flow (WGF) as follows:
\begin{definition}
Suppose we have a Wasserstein space $\mathcal{W}_2=(\mathcal{P}_2(\mathbb{R}^d),W_2)$, $\mathcal{P}_2(\mathbb{R}^d)=\{\mu \in \mathcal{P}_2,  \int ||\theta||^2 d\mu(\theta) < \infty \}$, $W_2(\mu_1,\mu_2)=min_{ \omega \in \Pi (\mu_1,\mu_2 )} \int {||\theta-\theta'||}^{2} d \omega(\theta,\theta')$.

A curve of $\mu_t$ is a Wasserstein Gradient Flow for functional $F$, if it satisfies Eq. \ref{eq:4}.
\begin{equation}
    \label{eq:4}
    \partial_t\mu_t=\nabla \cdot (\mu_t \nabla_\theta {\delta F(\mu_t) \over \delta \mu} )=\nabla \cdot (\mu_t \nabla_{W_2} F(\mu_t) )
\end{equation}
\end{definition}
WGF can be discretized as Stochastic Gradient Langevin Dynamics (SGLD) \cite{welling2011bayesian, chen2018unified} as Eq. \ref{eq:5}, where each particle $\theta$ follows the true posterior distribution $\pi$ with Gaussian perturbation. 
\begin{equation}
 \label{eq:5}
\theta_{t+1}^{i}=\theta_{t}^{i}-h(\nabla_{\theta_t^i} V(\theta^{i}_t))+\sqrt{2h} \epsilon, \epsilon \sim N(0,I)
\end{equation}
Whereas the Gaussian noise can assure the diversity of parameters, the learning can be unstable, when the learning rate is high.  

Hence, this paper relies on Stein Variational Gradient Descent (SVGD) \cite{NIPS2016_b3ba8f1b}, which is a version of Wasserstein Gradient Flow with Reproducing Kernel Hilbert Space (RKHS) \cite{chen2018unified}. In SVGD, interaction between particles $\theta_i$ guarantees their convergence to true posterior distribution with forcing diversity.
\begin{equation}
    \label{eq:6}
    \theta_{t+1}^i=\theta_{t}^i - {h \over M} \sum_{j=1}^M  K(\theta_t^i,\theta_t^j) \nabla_{\theta_t^j} V(\theta_t^j)-\nabla_{\theta_t^j}K(\theta_t^i,\theta_t^j)
\end{equation}
By using SVGD to approximate true posterior distribution, context vectors $\theta^j$ can be optimized to follow the true posterior distribution, effectively capturing a representation space of the image features.

\paragraph{Data-Dependent Prior}
In a Bayesian neural network, the prior is commonly chosen as the Standard normal distribution, i.e. BPL, which is data-independently initialized with zero means. Since such distribution does not include any information on data, it only regularizes the context vector in the neighbor of zero-mean. 
In many domains \cite{Li2020Data-dependent,lee2022priorgrad}, the data-dependent prior is utilized to improve the prior knowledge more informative. 

This paper utilizes the data-dependent prior for prompt learning, which is not restricted to the standard Gaussian distribution. Specifically, this paper derives the prior distribution to be dependent on image features, which can have multiple modes in their distributions.


\section{Method}
This section introduces adaptive particle-based prompt learning (APP) by enumerating the model formulation and by explaining its inference method. 
\subsection{Formulation of Prompt Posterior Distribution}
Following the CoOp formulation \cite{zhou2022learning}, we additionally reformulate the posterior distribution of context vector $\theta$ as Eq. \ref{eq:7}. 
\begin{align}
\label{eq:7}
    \pi(\theta)=p(\theta|X,Y)&= {p(Y|X,\theta )p(\theta|X)p(X) \over p(X,Y)} \notag \\&\propto p(Y|X,\theta )p(\theta|X)
\end{align} 
$\pi(\theta)$ is true posterior distribution, which is factorized with likelihood $p(Y|X,\theta )$ and conditioned prior $p(\theta|X) $. For likelihood Eq. \ref{eq:1}, we follow the formulation of CoOp, Eq. \ref{eq:1}, and we propose the image feature conditioned prior Eq. \ref{eq:9}, where mean is parametrized as $\phi$ and $\overline{\phi(f)} \coloneqq {1 \over N}\sum_{i=1}^{N}\phi(f(x_i))$. We set the standard deviation of prior $\sigma$ as a hyper-parameter.
\begin{equation}
\label{eq:9}
\log p(\theta|X) \propto  -{(\theta-\overline{\phi(f)})^T(\theta-\overline{\phi(f)}) \over \sigma^2 }
\end{equation}

\paragraph{Bayesian Adaptation of Prompt to Test data}
In few-shot learning, there is uncertainty on whether the test data will follow the distribution of training data or not. If there is a difference between the two data distributions, such a difference will harm the generalization ability of VLP model. To model the uncertainty from the mismatch between training and test datasets in the few-shot learning framework, we reformulate the posterior distribution to consider the uncertainty of a test image $x'$. Since training and test datasets are i.i.d sample; and because the prior is assumed to follow the Gaussian distribution; we derive the posterior distribution as Eq.\ref{eq:reform}. 
\begin{align}
\label{eq:reform}
    \pi(\theta)=p(\theta|X,Y,x')&= {p(Y|X,\theta )p(\theta|X)p(\theta|x')p(X) \over p(X,Y)} \notag \\&\propto \underbrace{p(Y|X,\theta )p(\theta|X)}_{Training}\underbrace{p(\theta|x')}_{Testing}
\end{align} 
After approximating the Eq.\ref{eq:7}, we adapt the context vector $\theta$ with test data-dependent prior.

\subsection{Variational Inference for Prompt Posterior}
Since Eq. \ref{eq:7} is not tractable, we approximate the posterior distribution of $\pi$, using particle-based variational inference. Suppose that $q$ is a probabilistic measure of variational distribution, which generates the context vector $\theta$, in Wasserstein space. Eq. \ref{eq:11} defines the optimization problem to approximate the model posterior distribution by the variational distribution.
\begin{equation}
\label{eq:10}
V(\theta) \coloneqq -\log p(Y|X,\theta ) - \log p(\theta|X)
\end{equation}
\begin{equation}
\label{eq:11}
    F(q) \coloneqq D_{KL}(q||\pi) \approx \E_{q} [V(\theta) + \log q]
\end{equation}
To define the steepest direction of Eq. \ref{eq:11}, we follow Wasserstein Gradient Flow Eq.\ref{eq:4}. By solving the Wasserstein Gradient Flow in Reproducing Kernel Hilbert Space (RKHS), we can define the following Wasserstein Gradient for variational distribution $q$, where the linear operator $\mathcal{K}_{q} T(\theta) \coloneqq \E_{\theta' \sim q }  [ K({\theta},{\theta'}) T({\theta'})]$. 
\begin{equation}
\label{eq:12}
    \partial_t q_t=\nabla \cdot (q_t \mathcal{K}_{q} \nabla_q {\delta F(q) \over \delta q} )
\end{equation}
Following the Continuity Equation, we can define the evolving path of $\theta_t \sim q_t$ as follows. 
\begin{align}
    \label{eq:13}
    \notag
    \partial_t \theta_t&=- \mathcal{K}_{q} \nabla_{\theta} ({\delta F(q) \over \delta q} ) \\ & = - [\int K({\theta},{\theta'}) \nabla_{\theta'} V(\theta')dq-\int \nabla_{\theta'} K(\theta,\theta')dq ]
\end{align}
By discretizing Eq. \ref{eq:13}, we derive the Stein Variational Gradient Descent \cite{NIPS2016_b3ba8f1b}, where each context vector $\theta^j$ can be optimized as follows.
\begin{equation}
    \label{eq:14}
    \theta_{t+1}^i=\theta_{t}^i - {h \over M} \sum_{j=1}^M  [K(\theta_t^i,\theta_t^j) \nabla_{\theta_t^j} V(\theta_t^j)-\nabla_{\theta_t^j}K(\theta_t^i,\theta_t^j)]
\end{equation}
For Eq. \ref{eq:14}, the first term can be interpreted as a smoothing gradient between context vectors $\theta^j$ and assure the convergence toward the true posterior distribution. The second term can be interpreted as the repulsive force between context vectors $\theta^j$ and guide the text features can cover the multi modes sparsely.
\subsection{Parameter Training of Data-Dependent Prior}
Since the prior has the parametrized mean $\phi$, we pre-train the $\phi$, which can map the image feature on the prompt space. To preserve the image feature information within our prior distribution, we propose to maximize the mutual information $I\left(\phi\left(f\left(x\right)\right);f(x)\right)$. In other words, this mutual information encourages the prior to capture the dependencies between the image features and the parameter of the prompt distribution. Due to the data processing inequality \cite{beaudry2012intuitive}, we derive inequality as follows: 
\begin{equation}
\label{eq:15}
     I(\phi(f(X));f(X)) \ge I(g(\phi(f(X)), \cdot);f(X))
\end{equation}
where $\phi$ can be learned to maximize the mutual information. 

\begin{prop}{Suppose that the Markov chain assumption holds as $f(X) \rightarrow \phi(f(X)) \rightarrow g(\phi(f(X)), \cdot) $, then the lower bound of the mutual information, $I(f(X);\phi(f(X)))$, is derived as follows: }
\label{prop:1}
\\ $I(f(X);\phi(f(X))) \ge I(f(X);g(\phi(f(X)), \cdot)) \ge \log C - \mathcal{L}_{CE} (\phi(f(X)),X,Y)$ 
\end{prop}
Based on Proposition \ref{prop:1}, we can maximize the mutual information by minimizing the cross entropy. The full training scenario is reported in Algorithm \ref{alg:prior}.

\subsection{Adaptation $\theta$ with Test Data-Dependent Prior}
Following the training of the posterior distribution as described in Eq. \ref{eq:7}, we extend our approach to accommodate an unseen data instance, $x'$, within the posterior distribution. This involves updating the context vector $\theta \sim q$ through a linear combination with the prior mean $\phi(f(x'))$. For the sake of simplicity, we perform a weighted average of the text features, which can be outlined as follows. The adaptation scenario is reported in Algorithm \ref{alg:test}.
\begin{equation}
    \label{eq:linearcomb}
     g(\theta^{*},y)=\alpha g(\theta,y)+(1-\alpha)g(\phi(f(x')),y)
\end{equation}

\begin{algorithm}[t!]
    \caption{Training Scenario of APP}
    \label{alg:prior}
\begin{algorithmic}[1]
   \State {\bfseries Input}: Dataset $\mathcal{D}=\{ X,Y \}$, Context vector $\theta^i$, Prior Network $\phi$ 
   
    \While{ not converged}
    \State Compute $\mathcal{L}_{CE} (\phi(f(X)),X,Y)$
    \State Update $\phi_{t+1}=\phi_{t}-h \nabla_{\phi} \mathcal{L}_{CE}$
    \EndWhile

\While{ not converged}
    \State Compute $V(\theta)= -\log p(Y|X,\theta ) - \log p(\theta|X) $ 
    \State Update $    \theta_{t+1}^i=\theta_{t}^i - {h \over M} \sum_{j=1}^M [K(\theta_t^i,\theta_t^j)\nabla_{\theta_t^j}V(\theta_t^j)-\nabla_{\theta_t^j}K(\theta_t^i,\theta_t^j)]$ $,\forall{i \in [1,...,M]}$
\EndWhile
\end{algorithmic}
\end{algorithm}

\begin{algorithm}[t!]
    \caption{Test Scenario of APP}
    \label{alg:test}
\begin{algorithmic}[1]
   \State {\bfseries Input}: Test image instance $x'$, Context vector $\theta^i$, Prior Network $\phi$ 
    \State Training as Algorithm \ref{alg:prior}
    \State Compute $g(\theta^{*},y_j)$ as Eq.\ref{eq:linearcomb}, $\forall{j \in [1,..,K]} $
    \State $y'=\text{argmax}_{y_j} \text{sim}(g(\theta^{*},y_j),f(x'))$

\end{algorithmic}
\end{algorithm}

\begin{figure*}[h!]
\centering
\begin{subfigure}[h!]{0.9\linewidth}
\centering
\includegraphics[width=.21\linewidth]{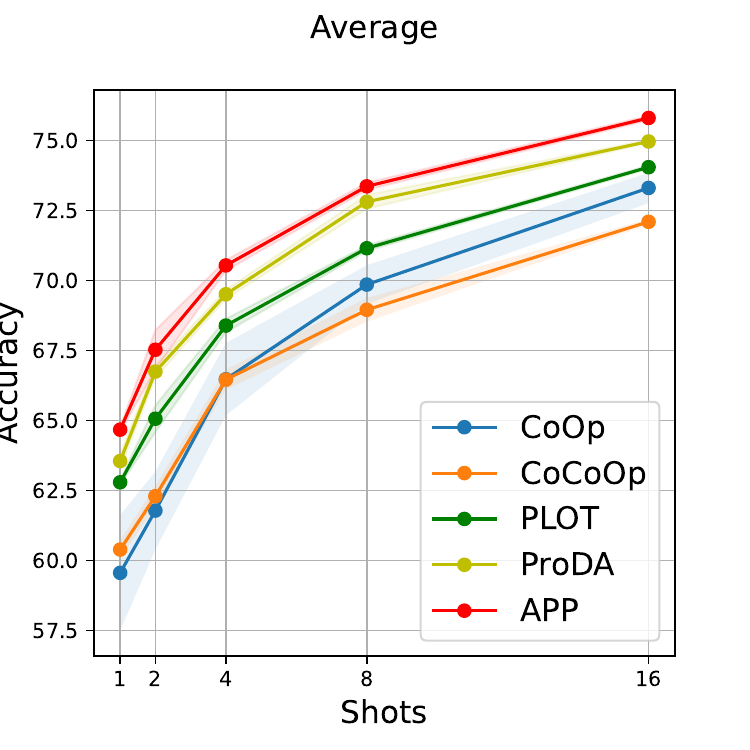}
\includegraphics[width=.21\linewidth]{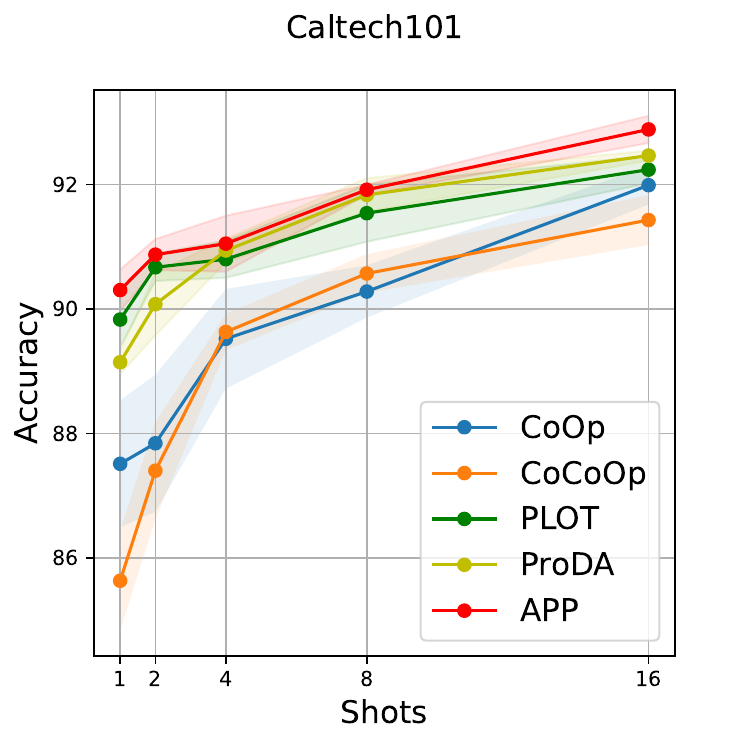}
\includegraphics[width=.21\linewidth]{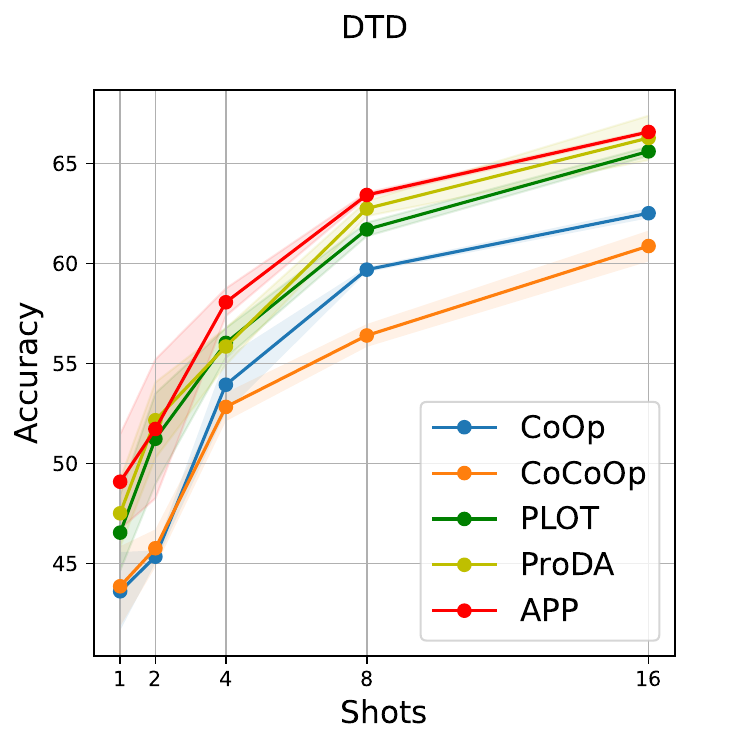}
\includegraphics[width=.21\linewidth]{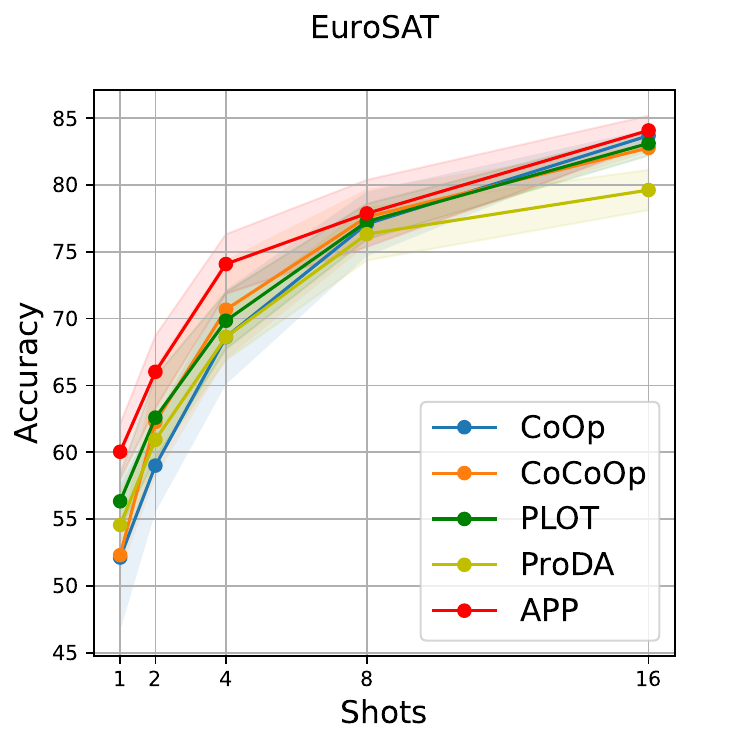}
\includegraphics[width=.21\linewidth]{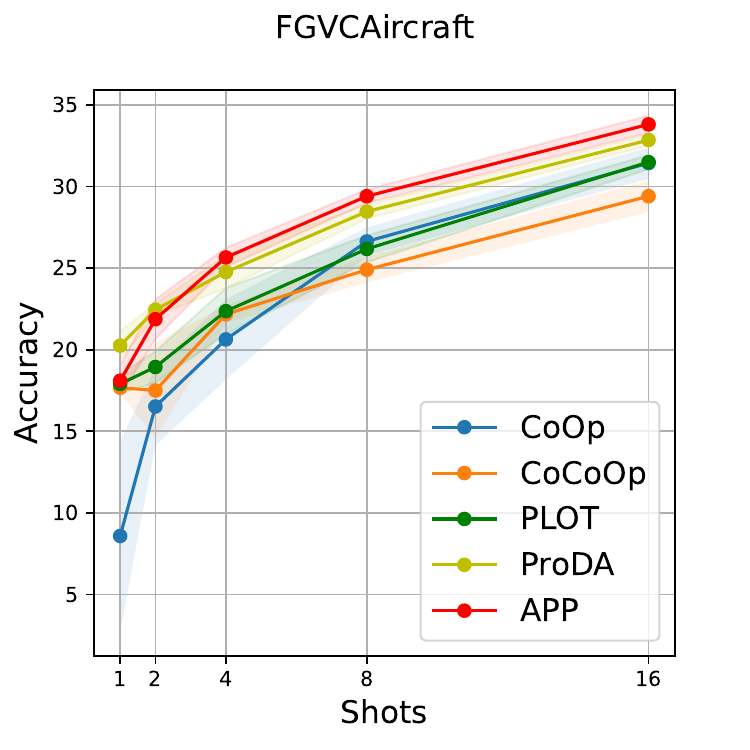}
\includegraphics[width=.21\linewidth]{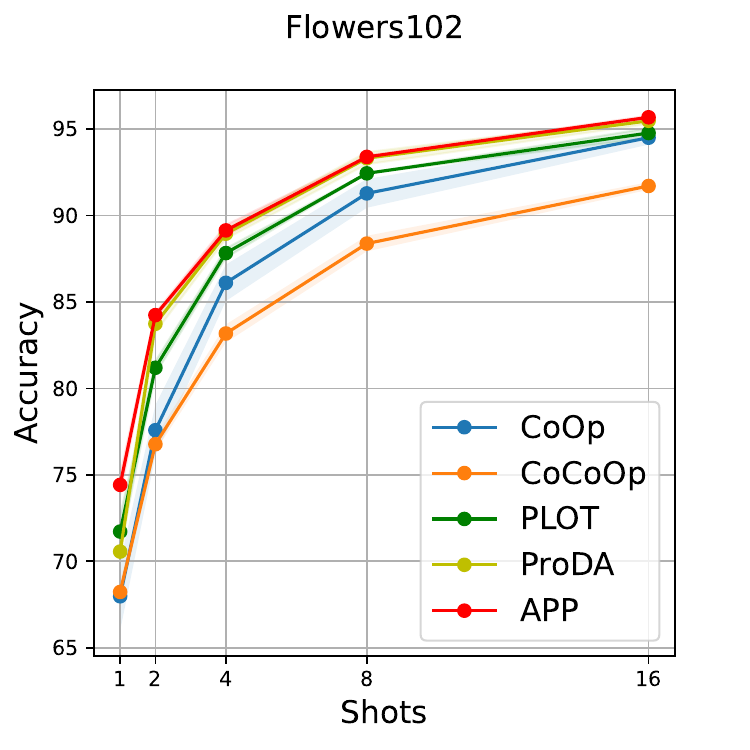}
\includegraphics[width=.21\linewidth]{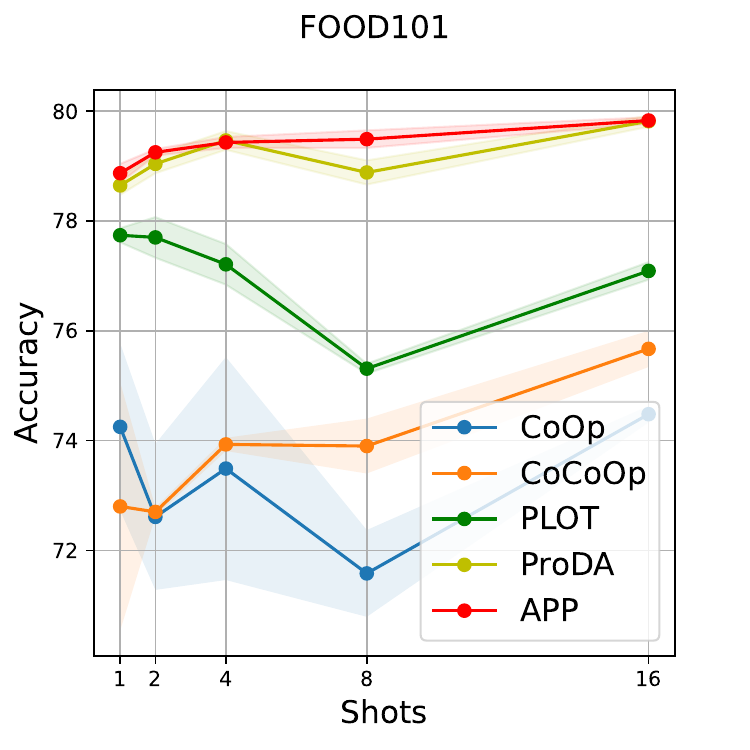}
\includegraphics[width=.21\linewidth]{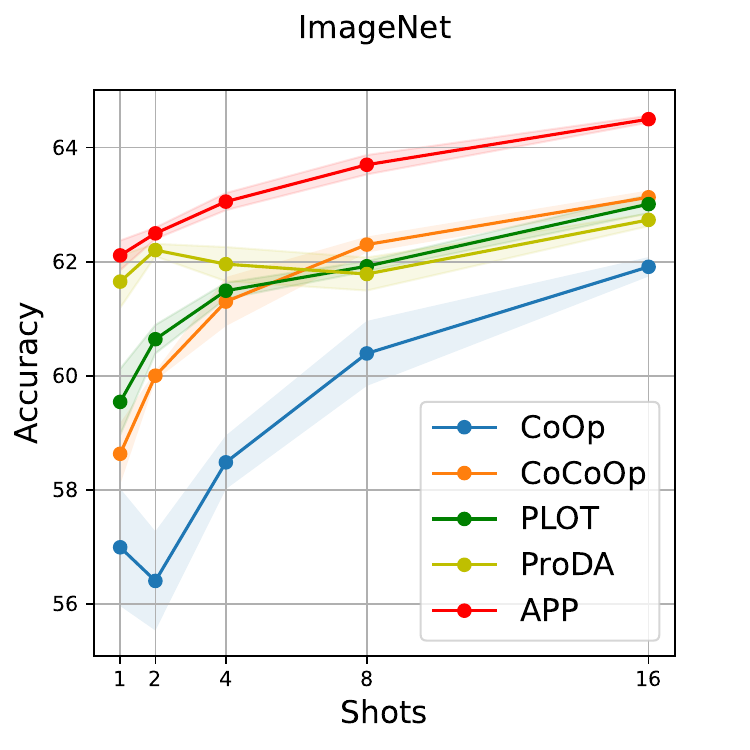}
\includegraphics[width=.21\linewidth]{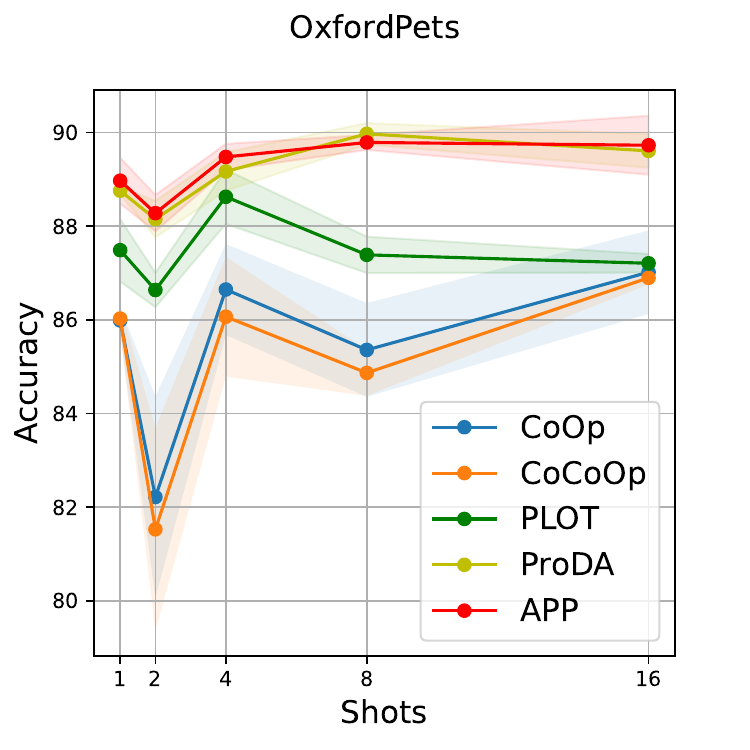}
\includegraphics[width=.21\linewidth]{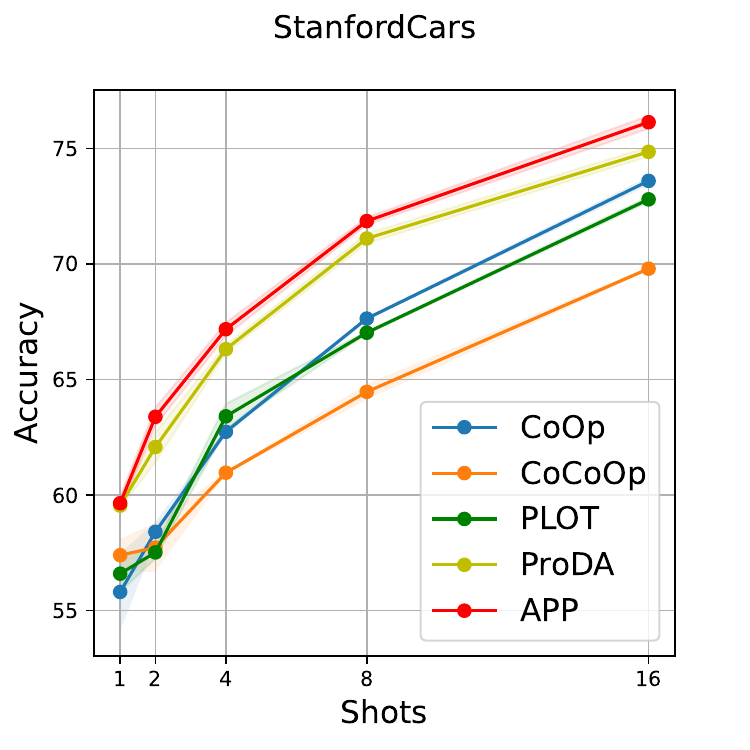}
\includegraphics[width=.21\linewidth]{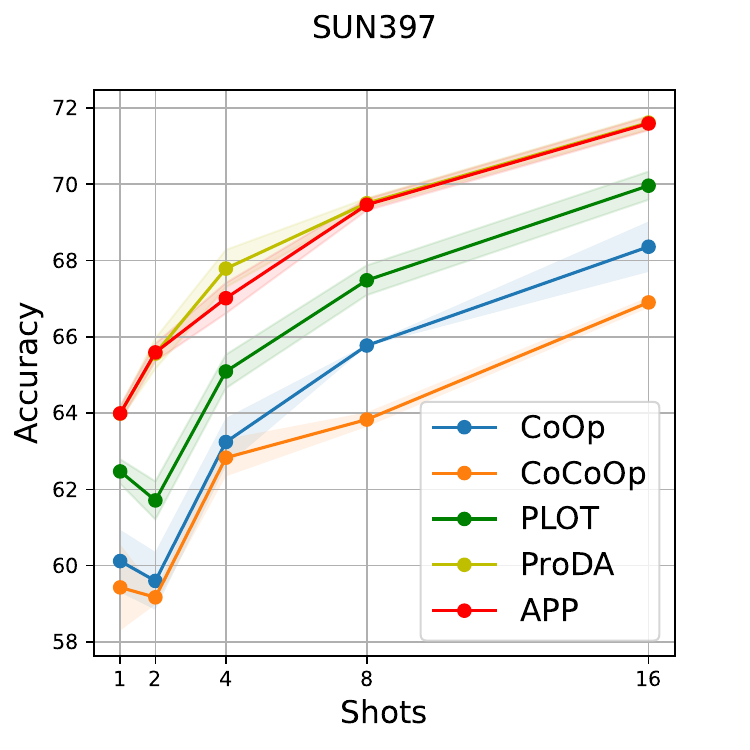}
\includegraphics[width=.21\linewidth]{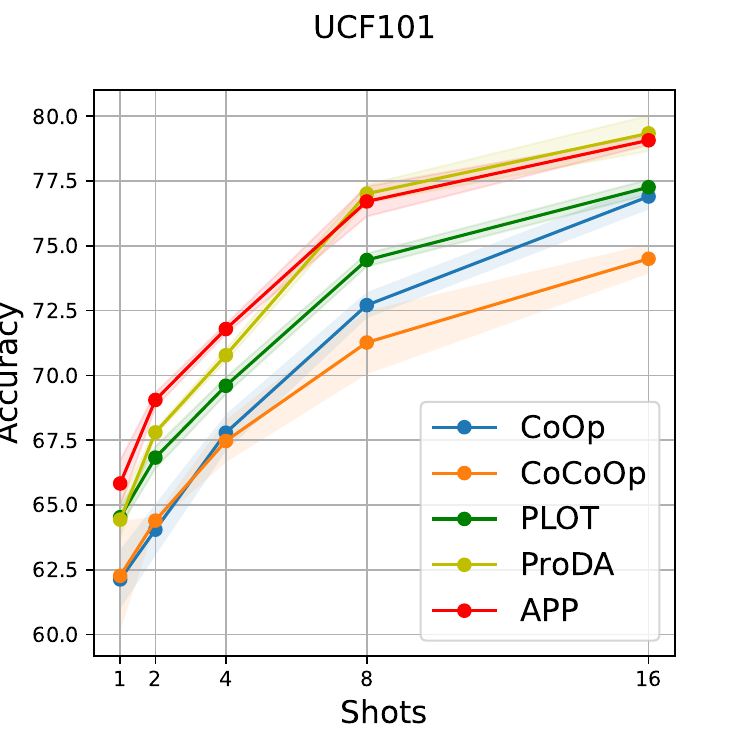}    
\end{subfigure}
\vskip -0.1in
\caption{Result of Few-shot Classification. We conduct three-replicated experiments. }
\label{fig:fewshot}
\vskip -0.2in
\end{figure*}
\section{Results}
\subsection{Experiment Settings}
We conduct three distinct experiments; Few-shot classification, domain generalization of ImageNet, and base-to-new generalization following PLOT \cite{chen2023plot}. For Few-shot classification, we conduct 11 image datasets, including Caltech101 \cite{fei2004learning}, DTD \cite{cimpoi2014describing}, EuroSAT \cite{helber2019eurosat}, FGVCAircraft \cite{maji2013fine}, Oxford 102 Flower \cite{nilsback2008automated}, OxfordPets \cite{6248092}, Food101 \cite{bossard2014food}, StanfordCars \cite{krause20133d}, Sun397 \cite{xiao2010sun}, UCF101 \cite{soomro2012ucf101}, and ImageNet \cite{deng2009imagenet}. We followed the training setting of PLOT \cite{chen2023plot}, where the training shots are chosen in 1, 2, 4, 8, 16 shots, and we train 50, 100, 100, 200, and 200 epochs for each shot. For ImageNet, we train the prompts in 50 epochs for all shots. Before the training context vector $\theta$, we train the prior mean $\phi$ in 10, 20, 20, 40, and 40 epochs for each shot.
For domain generalization of ImageNet, we train prompts about ImageNet as a source dataset and report the accuracy of the source dataset and target datasets, including ImageNetV2 \cite{recht2019imagenet}, ImageNet-A \cite{hendrycks2021natural}, ImageNet-R \cite{hendrycks2021many}, and ImageNet-Sketch \cite{wang2019learning}. For base-to-new generalization, we train prompts using 16 shots for each of 11 datasets for the base class and report the performance of base and new classes. 

As a common setting, we conduct three replicated experiments to report the performances, and we use CLIP \cite{jia2021scaling} as the backbone network, where ResNet50 \cite{he2016deep} is chosen as the image encoder. All context vectors are vectors of 16 dimensions, which are sampled from $N(0,0.02I)$, and class information is inserted at the end position of context vectors. We fixed the precision $1/ \sigma^2$ as 1.0 for all settings, and RBF kernel is used for $K$. The adaptation weight $\alpha$ is chosen as 0.9 for few-shot classification, and 0.7 for generalization experiments. We report more details of the setting in the Appendix.

\paragraph{Baselines} We compare the performance of our method, APP with CoOp \cite{zhou2022learning}, CoCoOP \cite{zhou2022conditional}, PLOT \cite{chen2023plot}, and ProDA \cite{lu2022prompt}. We do not include BPL \cite{derakhshani2023bayesian} as our baselines due to reasons in the Appendix. We initialize the four context vectors for PLOT, ProDA, and APP randomly.
\subsection{Few-Shot Classification}
\paragraph{Quantitative Analysis}
Figure \ref{fig:fewshot} indicates that our method outperforms baselines on all benchmark datasets on average, where our performance is superior to 45 out of 55 experiment cases (11 datasets $\times$ 5 shots). The advantage of APP stands out in Caltech101, DTD, EuroSAT, and ImageNet, which consist of more diverse images. Since the data-dependent prior and the repulsive force of Eq. \ref{eq:14} enable text features to infer the multi-modes of image features, our context vectors are learned to capture the diverse semantics of image features.

\paragraph{Qualitative Analysis}
\begin{figure}[h!]
\vskip -0.1in
\centering
\begin{subfigure}[h]{0.15\textwidth}
\includegraphics[width=\textwidth]{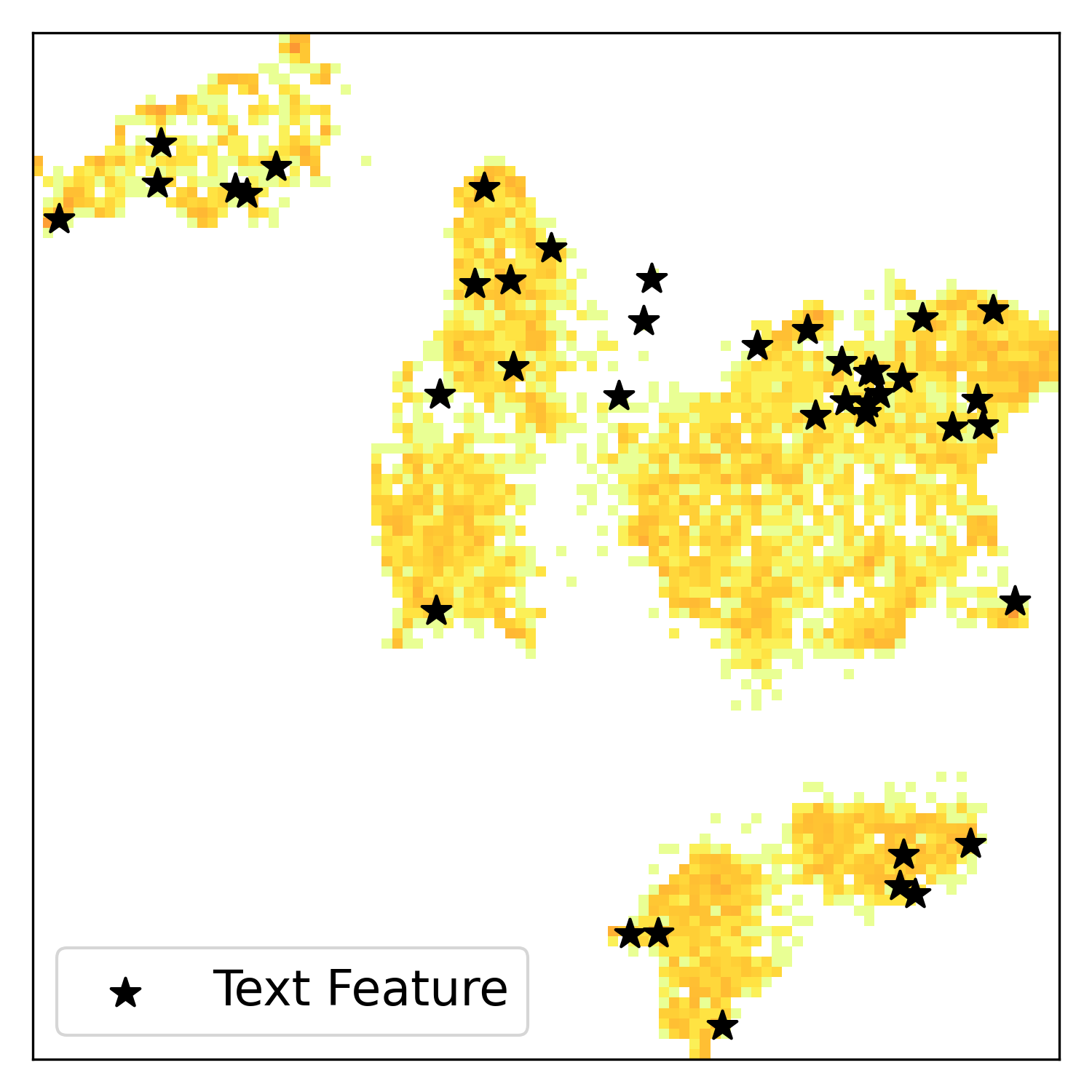}
\end{subfigure}
\hfill
\begin{subfigure}[h]{0.15\textwidth}
\includegraphics[width=\textwidth]{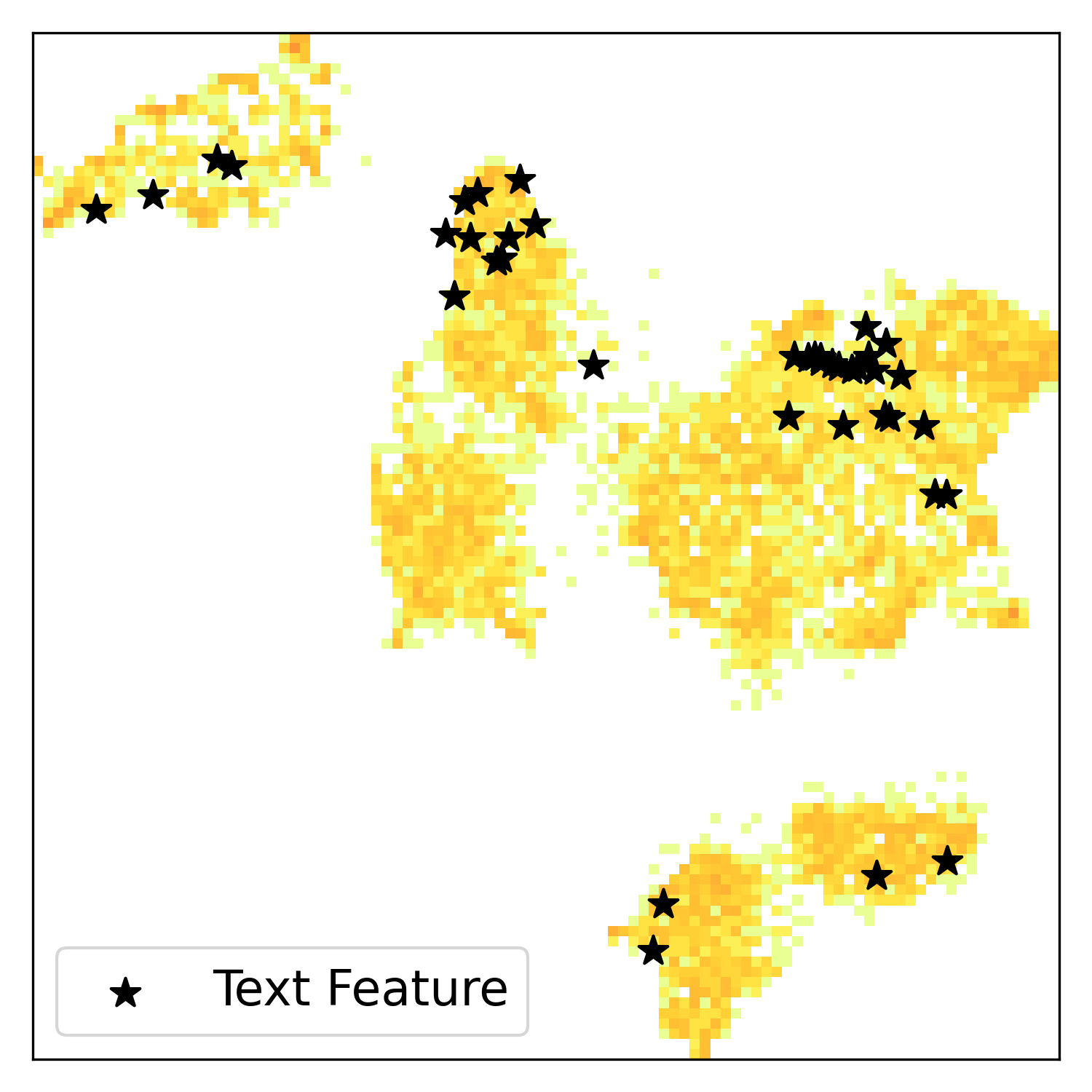}
\end{subfigure}
\hfill
\begin{subfigure}[h]{0.15\textwidth}
\includegraphics[width=\textwidth]{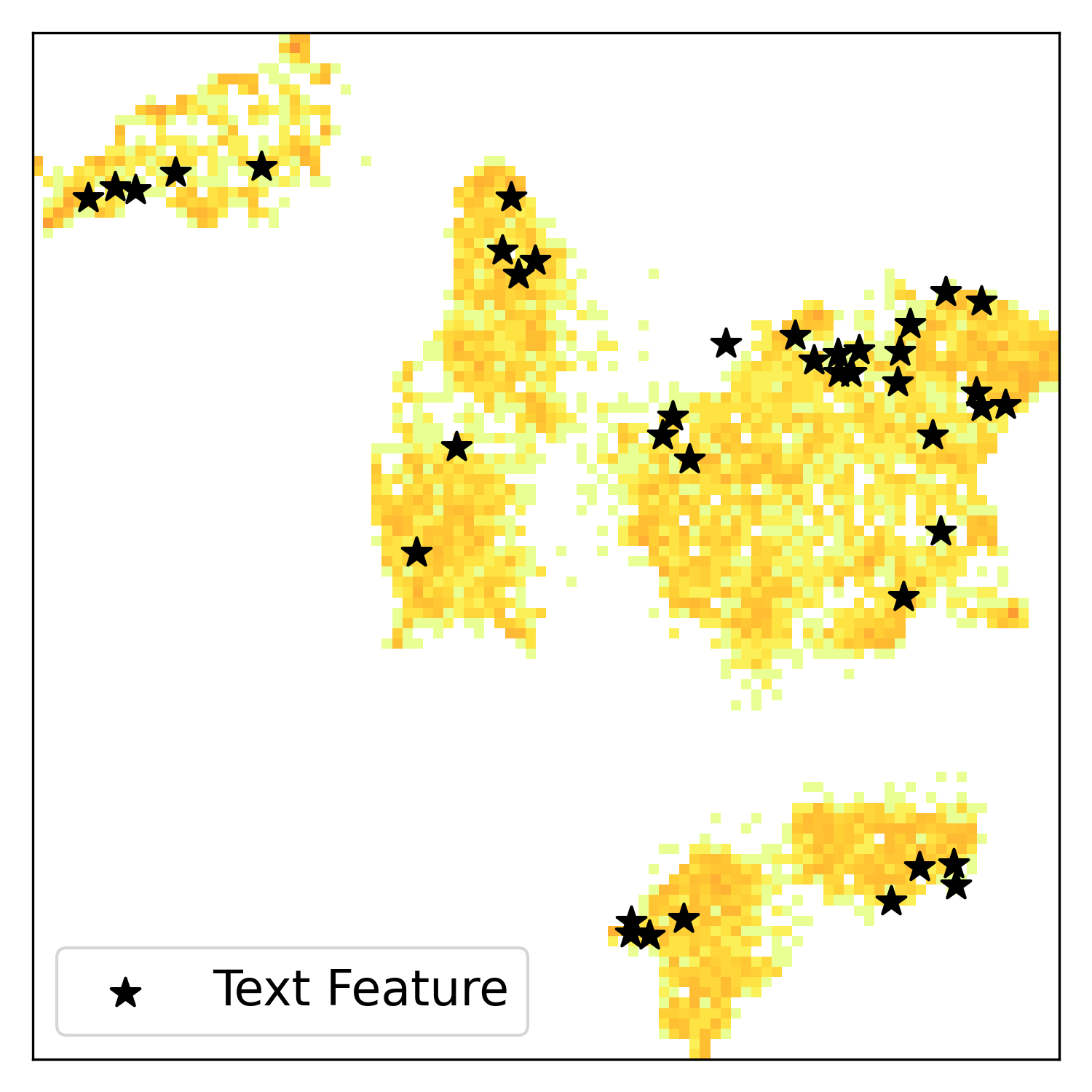}
\end{subfigure}

\begin{subfigure}[h]{0.15\textwidth}
\includegraphics[width=\textwidth]{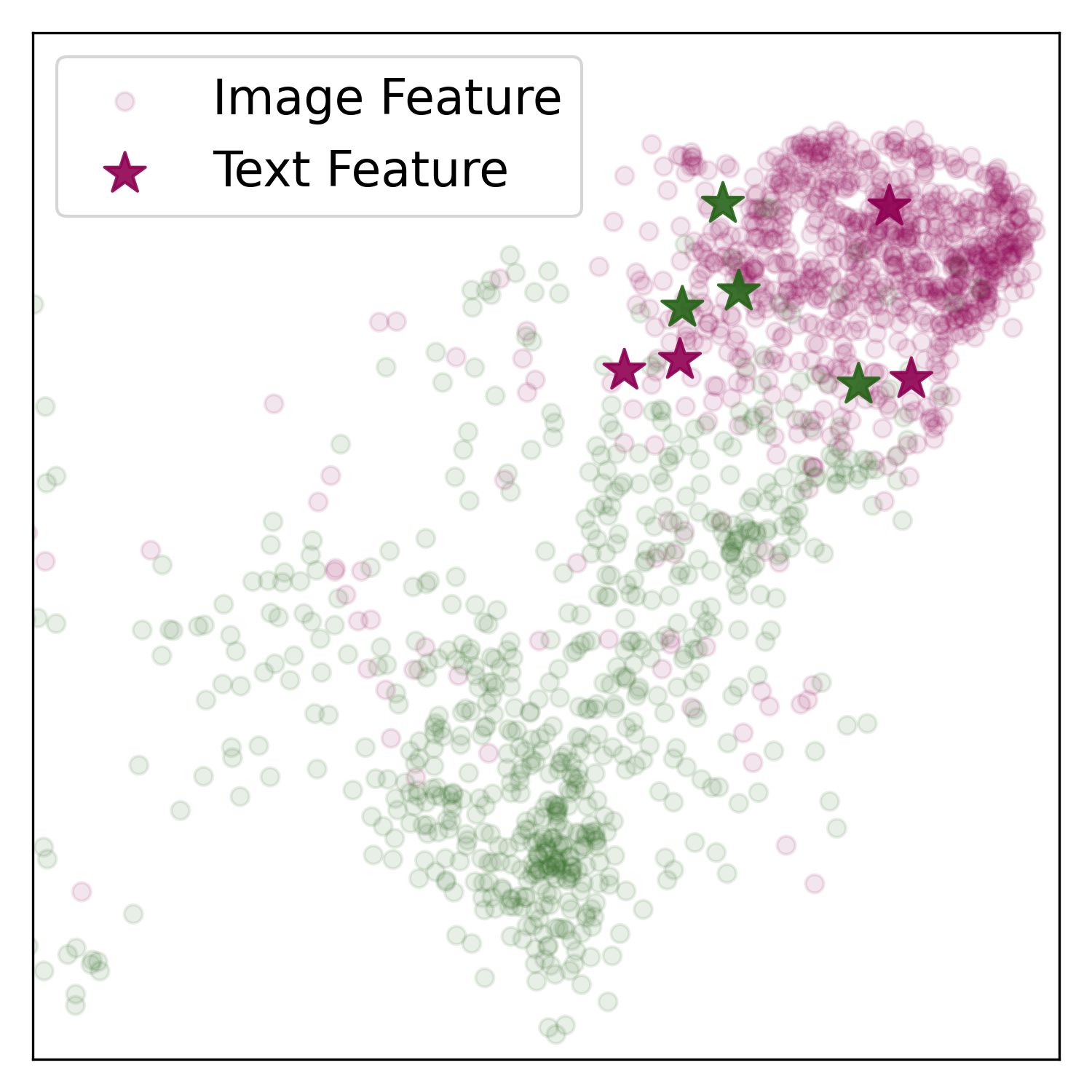}
\caption{PLOT}
\end{subfigure}
\hfill
\begin{subfigure}[h]{0.15\textwidth}
\includegraphics[width=\textwidth]{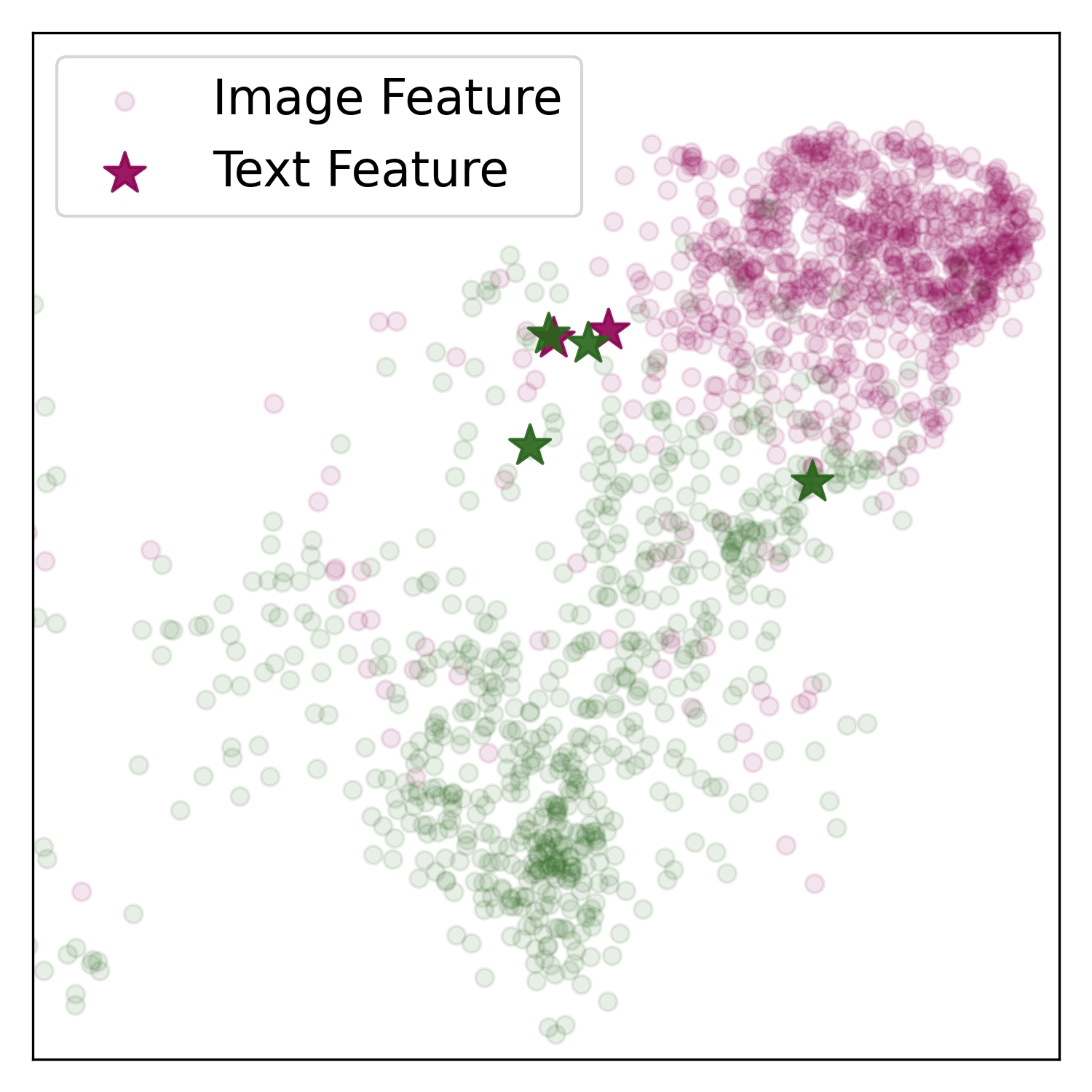}
\caption{ProDA}
\end{subfigure}
\hfill
\begin{subfigure}[h]{0.15\textwidth}
\includegraphics[width=\textwidth]{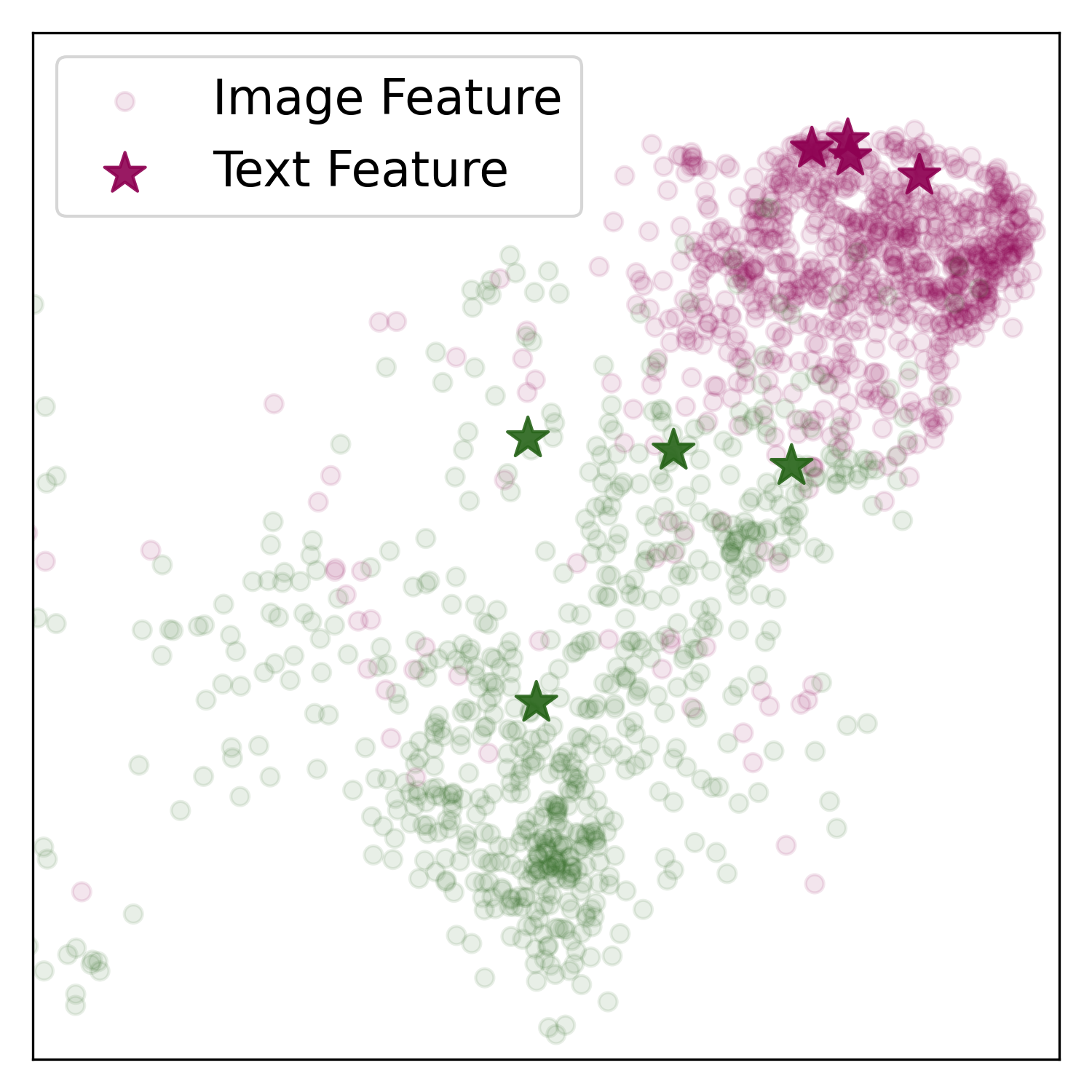}
\caption{APP}
\end{subfigure}
\caption{Umap visualization about image features and text features for EuroSAT. \textbf{(Upper)} Histograms correspond to image features and $\star$ means text features of all classes. \textbf{(Lower)} Image and text features of arbitrary two classes. The color coding corresponds to each class.}
\label{fig:umap}
\vskip-0.2in
\end{figure}
To demonstrate the efficacy of our method in capturing intricate image features, we provide visualizations of both image features ($f(x_i)$) and text features ($g(\theta_j,y_i)$). In Figure \ref{fig:umap}, we present Umap \cite{mcinnes2018umap} representations for EuroSAT test dataset. The upper figures depict image and text representations of all classes. A richer yellow hue indicates a denser allocation of image features. Our method, APP, well captures and comprehensively spans various modes within image features. The lower figures spotlight two arbitrarily selected classes, revealing how APP's text features harmoniously align with the image features of each class. This alignment is particularly pronounced in comparison to other methods.

Table \ref{tab: kmeans variance} provides the numerical analysis of the alignment between text feature representations from our method and image feature modes. We fitted k-means clustering on the test image features with k as a hyperparameter. Then, we counted the number of prompts assigned on each cluster, which will be similar for all clusters if prompts are adequately assigned to several modes of image features. In this context, we present the variance of prompt counts and show that text prompt representations of APP are well distributed across image feature representations.
\begin{table}[h]
\centering
\vskip-0.05in
\caption{Variance of the number of text feature representations assigned to each cluster after fitting k-means clustering to image features. \textbf{Bold} means the smallest of each column.}
\resizebox{.8\columnwidth}{!}{%
\begin{tabular}{c|cccccc} \toprule
\multirow{2}{*}{\textbf{Methods}}&\multicolumn{6}{c}{\textbf{k} (Number of clusters)}\\ \cmidrule(lr){2-7}
 & 5 & 6 & 7 & 8 & 9 & 10 \\ \midrule
PLOT & 26.0 & 27.6 & 11.9 & 7.5 & 8.5 & 7.8 \\
ProDA & 53.2 & 26.6 & 25.3 & 12.0 & 9.8 & 9.0 \\
APP & \textbf{25.2} & \textbf{21.9} & \textbf{9.9} & \textbf{6.3} & \textbf{4.0} & \textbf{2.6} \\ \bottomrule
\end{tabular}%
}
\label{tab: kmeans variance}
\vskip-0.1in
\end{table}

\paragraph{Ablation Study}
We show two ablation studies, considering our method and the number of prompts.

To identify the key enabler, we conduct additional ablation studies for APP by experimenting on 1) data-dependent prior and 2) Stein Variational Gradient Descent (SVGD). For all cases, four context vectors are initialized, and we select MLE training as the baseline optimized by SGD. Table \ref{tab:test_ablation} shows the posterior approximation with data-dependent prior improved performances generally than MLE Training. 
SVGD shows a more robust performance in a few data instances than SGD. The posterior approximation by SVGD also shows consistently better performance, outperforming in Caltech101 and EuroSAT dataset. 
\begin{table}[h!]
\caption{An ablation study about our method, APP. Experiments are replicated over three times.}
\label{tab:test_ablation}
\centering
\resizebox{\columnwidth}{!}{
\begin{tabular}{c|c|c|c|c} \toprule
\multirow{2}{*}{\textbf{Dataset}}&\multirow{2}{*}{\textbf{Methods}}&\multicolumn{3}{c}{\textbf{Number of shots}}\\ \cmidrule(lr){3-5}
&&\textbf{1}&\textbf{2}&\textbf{4}\\ \midrule
\multirow{4}{*}{Caltech101} & SGD    & $89.87 \pm 0.02 $  & $90.29 \pm 0.20 $  & $91.00 \pm 0.29 $  \\ 
& SVGD & $89.90 \pm 0.07 $&  $90.56 \pm 0.10 $& $91.02 \pm 0.29 $  \\
 & SGD+Prior  & $90.26 \pm 0.21$  & $90.70 \pm 0.25 $  & $91.01 \pm 0.14 $  \\
  & SVGD+Prior (APP)      & $\mathbf{90.30 \pm 0.33} $  & $ \mathbf{90.87 \pm 0.25}$  & $\mathbf{91.05 \pm 0.45}$  \\ \midrule
\multirow{4}{*}{EuroSAT} & SGD    & $56.43 \pm 2.31 $  & $63.93 \pm 2.85 $  & $72.63 \pm 2.74 $   \\
& SVGD & $58.93 \pm 4.28 $&  $64.36 \pm 3.16 $& $73.64 \pm 2.00 $ \\
 & SGD+Prior          & $59.96 \pm 4.77$  & $63.99 \pm 2.88 $  & $72.63 \pm 2.76 $  \\
  & SVGD+Prior (APP)      & $\mathbf{60.04 \pm 2.08} $  & $\mathbf{66.02 \pm 2.70}$  & $\mathbf{74.08 \pm 2.23}$\\ \midrule
\multirow{4}{*}{Food101} & SGD    & $78.53\pm 0.14 $  & $78.88 \pm 0.09 $  & $78.88 \pm 0.25 $    \\
& SVGD & $78.51 \pm 0.12 $&  $78.86 \pm 0.07 $& $78.87 \pm 0.26 $\\
 & SGD+Prior          & $78.81 \pm 0.16$  & $79.15 \pm 0.07 $  & $79.32 \pm 0.08$   \\
  & SVGD+Prior (APP)      & $\mathbf{78.87 \pm 0.17}$  & $ \mathbf{79.25\pm 0.06}$  & $\mathbf{79.43 \pm 0.10}$   \\ \bottomrule
\end{tabular}
}
\vskip -0.1in
\end{table}
We also carry out additional experiments varying the number of prompts to explore the influence of samples within the posterior distribution as table \ref{tab:numablation}. While there is a tendency for higher performance with an increased number of prompts, please note that the performance can sufficiently be achieved with approximately four prompts.

\begin{table}[h!]
\caption{An ablation study with regard to the number of prompts (\textbf{M}). Experiments are replicated over three times.}
\label{tab:numablation}
\centering
\resizebox{\columnwidth}{!}{
\begin{tabular}{c|c|c|c|c} \toprule
\multirow{2}{*}{\textbf{Dataset}}&\multirow{2}{*}{\textbf{M}}&\multicolumn{3}{c}{\textbf{Number of shots}}\\ \cmidrule(lr){3-5}
& & \textbf{1} &\textbf{2} & \textbf{4} \\ \midrule
\multirow{3}{*}{Caltech101} & 2    & $	89.19 \pm 0.20 $  & $	90.08 \pm 0.22 $  & $	90.67 \pm 0.69 $  \\ 
& 4 & $\mathbf{90.30 \pm 0.33} $  & $ \mathbf{90.87 \pm 0.25}$  & $91.05 \pm 0.45$  \\
 & 8  & $90.18 \pm 0.28$  & $	90.33 \pm 0.29 $  & $\mathbf{91.38 \pm 0.16} $  \\ \midrule
\multirow{3}{*}{EuroSAT} & 2  & $52.06 \pm 4.98 $  & $60.65 \pm 3.00 $  & $71.19 \pm 1.78 $   \\
&4 & $\mathbf{{60.04 \pm 2.08}} $  & $ \mathbf{66.02 \pm 2.70}$  & $74.08 \pm 2.23$ \\
 & 8  & $	58.60 \pm 2.35$  & $	64.22 \pm 5.32 $  & $\mathbf{	74.32 \pm 1.97}$  \\ \midrule
\multirow{3}{*}{Food101} & 2   & $	78.27\pm 0.16 $  & $78.74 \pm 0.09$  & $78.93 \pm 0.12 $    \\
& 4&  $78.87 \pm 0.17$  & $79.25\pm 0.06$  & $79.43 \pm 0.10$  \\
 &8 & $\mathbf{78.90 \pm 0.21}$  & $\mathbf{79.29 \pm 0.09} $  & $\mathbf{79.45 \pm 0.12} $   \\ \bottomrule
\end{tabular}
}
\vskip -0.1in
\end{table}

\paragraph{Time and Memory Complexity Comparison}
\label{sec:time}
In this paragraph, we compare the time and memory complexity between APP and CoCoOp. This comparison aims for showing the appropriateness of prompt learning in the context of efficiency. While both data-dependent prior and conditioned prompts share similarities in Figure \ref{fig:subfig1}, Table \ref{tab:time complexity} underscores the fact that data-dependent priors exhibit efficiency and effectiveness. This is because the regularization process of the data-dependent prior does not require gradient updates.
\begin{table}[h]
\centering
\caption{Time complexity for Caltech101. M means the number of prompts. $-$ is a training failure due to the memory issue. \textbf{BS} means Batch Size and \textbf{Acc} represents accuracy.}
\vskip-0.05in
\label{tab:time complexity}
\resizebox{0.9\columnwidth}{!}{
\begin{tabular}{c|c|c|c|c} \toprule
 \textbf{Methods} & \textbf{BS} & \textbf{Memory (GB)} &\textbf{Time (s)} & \textbf{Acc ($\%$)} \\ \midrule
CoCoOp & 10 & 20&289 &84.4
\\
(M=1) & 128& - &- &-
\\ \midrule
APP& 10 & 7.2&260 & 89.3  \\
(M=4)& 128 & 9.6& 112 & 90.2
\\ \bottomrule
\end{tabular}    
}
\vskip -0.1in
\end{table} 

\begin{table*}[h!]
\caption{Test accuracies ($\%$) of the unseen classes generalization settings. \textbf{H} means the harmonic mean between the base accuracy and the new accuracy. \textbf{Bold} means the best accuracy of each column. We report the mean accuracy with three times replications due to the space issue.}
\label{tab:base2}
\centering
\begin{minipage}{.22\linewidth}
\resizebox{\textwidth}{!}{
    \begin{tabular}{c|c c|c} \toprule
    \centering
    Method & Base & New & \bf{H}\\ \midrule
    CoCoOp & $71.7$ & $53.6$ & $61.4$ \\
    PLOT & $82.2$ & $60.5$ & $69.7$ \\
    ProDA & $82.4$ & $63.6$ & $71.8$ \\
    APP & $\mathbf{83.0}$ & $\mathbf{65.8}$ & $\mathbf{73.4}$ \\ \bottomrule
    \end{tabular}
    }
    \subcaption{Average}
\end{minipage} 
\begin{minipage}{.22\linewidth}
\centering
    \resizebox{\textwidth}{!}{
    \begin{tabular}{c|c c|c}\toprule
    \centering
    Method & Base & New & \bf{H}\\\midrule
    CoCoOp & $\mathbf{95.2}$ & $87.4$ & $91.2$ \\
    PLOT & $94.7$ & $88.1$ & $91.3$ \\
    ProDA & $\mathbf{95.2}$ & $86.8$ & $90.8$ \\
    APP & $\mathbf{95.2}$ & $\mathbf{91.0}$ & $\mathbf{93.0}$ \\ \bottomrule
    \end{tabular}
    }
    \subcaption{Caltech101}
\end{minipage}
\begin{minipage}{.22\linewidth}
    \resizebox{\textwidth}{!}{
    \begin{tabular}{c|c c|c}\toprule
    \centering
    Method & Base & New & \bf{H}\\\midrule
    CoCoOp & $74.6$ & $38.9$ & $51.1$ \\
    PLOT & $78.1$ & $42.7$ & $55.2$ \\
    ProDA & $78.0$ & $47.0$ & $58.6$ \\
    APP & $\mathbf{78.4}$ & $\mathbf{48.9}$ & $\mathbf{60.2}$ \\\bottomrule
    \end{tabular}
    }
    \subcaption{DTD}
\end{minipage}
\begin{minipage}{.22\linewidth}
\resizebox{\textwidth}{!}{
    \begin{tabular}{c|c c|c}\toprule
    \centering
    Method & Base & New & \bf{H}\\\midrule
    CoCoOp & $91.4$ & $35.6$ & $51.3$ \\
    PLOT & $92.9$ & $39.3$ & $55.2$ \\
    ProDA & $89.6$ & $39.0$ & $54.4$ \\
    APP & $\mathbf{93.6}$ & $\mathbf{47.6}$ & $\mathbf{63.1}$ \\\bottomrule
    \end{tabular}
    }
    \subcaption{EuroSAT}
\end{minipage}
\begin{minipage}{.22\linewidth}
\resizebox{\textwidth}{!}{
    \begin{tabular}{c|c c|c}\toprule
    \centering
    Method & Base & New & \bf{H}\\\midrule
    CoCoOp & $29.1$ & $14.1$ & $19.0$ \\
    PLOT & $43.3$ & $20.4$ & $27.8$ \\
    ProDA & $44.3$ & $24.1$ & $31.2$ \\
    APP & $\mathbf{44.9}$ & $\mathbf{26.0}$ & $\mathbf{33.0}$ \\\bottomrule
    \end{tabular}
    }
    \subcaption{FGVC-Aircraft}
\end{minipage}
\begin{minipage}{.22\linewidth}
\resizebox{\textwidth}{!}{
    \begin{tabular}{c|c c|c}\toprule
    \centering
    Method & Base & New & \bf{H}\\\midrule
    CoCoOp & $80.7$ & $78.8$ & $79.7$ \\
    PLOT & $83.4$ & $84.2$ & $83.8$ \\
    ProDA & $84.5$ & $\mathbf{86.2}$ & $\mathbf{85.4}$ \\
    APP & $\mathbf{84.6}$ & $86.1$ & $\mathbf{85.4}$ \\\bottomrule
    \end{tabular}
    }
    \subcaption{Food101}
\end{minipage}   
\begin{minipage}{.22\linewidth}
\resizebox{\textwidth}{!}{
    \begin{tabular}{c|c c|c}\toprule
    \centering
    Method & Base & New & \bf{H}\\\midrule
    CoCoOp & $68.3$ & $60.5$ & $64.1$ \\
    PLOT & $68.3$ & $58.4$ & $62.9$ \\
    ProDA & $68.8$ & $63.0$ & $65.7$ \\
    APP & $\mathbf{69.9}$ & $\mathbf{63.2}$ & $\mathbf{66.4}$ \\\bottomrule
    \end{tabular}
    }
    \subcaption{ImageNet}
\end{minipage}   
\begin{minipage}{.22\linewidth}
\resizebox{\textwidth}{!}{
    \begin{tabular}{c|c c|c}\toprule
    \centering
    Method & Base & New & \bf{H}\\\midrule
    CoCoOp & $94.7$ & $58.6$ & $72.4$ \\
    PLOT & $\mathbf{97.4}$ & $54.2$ & $69.6$ \\
    ProDA & $97.0$ & $58.5$ & $73.0$ \\
    APP & $96.8$ & $\mathbf{61.0}$ & $\mathbf{74.8}$ \\\bottomrule
    \end{tabular}
    }
    \subcaption{Flower102}
\end{minipage} 
\begin{minipage}{.22\linewidth}
\resizebox{\textwidth}{!}{
    \begin{tabular}{c|c c|c}\toprule
    \centering
    Method & Base & New & \bf{H}\\\midrule
    CoCoOp & $89.4$ & $\mathbf{91.0}$ & $90.2$ \\
    PLOT & $95.9$ & $87.6$ & $91.5$ \\
    ProDA & $96.4$ & $88.6$ & $\mathbf{92.4}$ \\
    APP & $\mathbf{96.8}$ & $88.3$ & $\mathbf{92.4}$ \\\bottomrule
    \end{tabular}
    }
    \subcaption{Oxford Pets}
\end{minipage} 
\begin{minipage}{.22\linewidth}
\resizebox{\textwidth}{!}{
    \begin{tabular}{c|c c|c}\toprule
    \centering
    Method & Base & New & \bf{H}\\\midrule
    CoCoOp & $68.7$ & $51.6$ & $58.9$ \\
    PLOT & $84.2$ & $62.6$ & $71.8$ \\
    ProDA & $84.5$ & $68.1$ & $75.5$ \\
    APP & $\mathbf{85.9}$ & $\mathbf{69.5}$ & $\mathbf{76.8}$ \\\bottomrule
    \end{tabular}
    }
    \subcaption{Stanford Cars}
\end{minipage} 
\begin{minipage}{.22\linewidth}
    \centering
\resizebox{\textwidth}{!}{
    \begin{tabular}{c|c c|c}\toprule
    Method & Base & New & \bf{H}\\\midrule
    CoCoOp & $73.3$ & $64.0$ & $68.4$ \\
    PLOT & $79.8$ & $65.3$ & $71.8$ \\
    ProDA & $\mathbf{80.9}$ & $70.8$ & $75.5$ \\
    APP & $80.6$ & $\mathbf{73.3}$ & $\mathbf{76.8}$ \\\bottomrule
    \end{tabular}
    }
    \subcaption{Sun397}
\end{minipage} 
\begin{minipage}{.22\linewidth}
    \centering
    \resizebox{\textwidth}{!}{
    \begin{tabular}{c|c c|c}\toprule
    Method & Base & New & \bf{H}\\\midrule
    CoCoOp & $79.2$ & $47.0$ & $59.0$ \\
    PLOT & $86.5$ & $62.7$ & $72.7$ \\
    ProDA & $\textbf{86.9}$ & $67.9$ & $76.2$ \\
    APP & $86.2$ & $\mathbf{69.2}$ & $\mathbf{76.8}$ \\\bottomrule
    \end{tabular}
    }
    \subcaption{UCF101}
\end{minipage} 
\vskip -0.2in
\end{table*}

\subsection{Generalization Experiment}
It is well known that VLP model is rather robust for domain shift \cite{radford2021learning}, yet this good property can be corrupted when the model parameter is fine-tuned on the downstream task \cite{wortsman2022robust}. Therefore, if this robustness regarding domain shift from VLP models could be sustained with prompt learning, it implies that this technique can be utilized more generally. For comparing the robustness, we conduct two experiments: 1) Unseen classes generalization setting in 11 datasets. and 2) Domain generalization setting in ImageNet. 

\paragraph{Unseen classes Generalization in 11 Datasets.}
Following \citet{zhou2022conditional}, we report the robustness over unseen classes in 11 datasets. Table \ref{tab:base2} shows the test accuracies with regard to both seen (base) classes and unseen (new) classes. Note that APP is robust to unseen (new) class data, maintaining the performance of seen class data, while other baselines have a performance trade-off between seen and unseen classes.

\begin{figure}[h!]
\vskip-0.1in
\centering
\begin{subfigure}[h]{0.32\columnwidth}
\includegraphics[width=\linewidth]{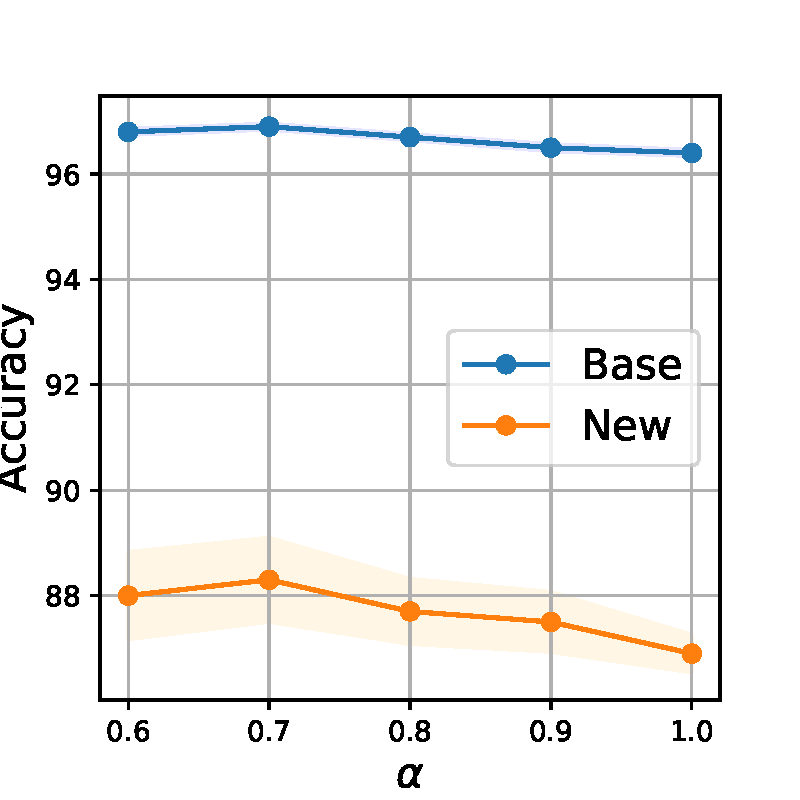}
\caption{Oxford Pets}
\end{subfigure}
\hfill
\begin{subfigure}[h]{0.32\columnwidth}
\includegraphics[width=\linewidth]{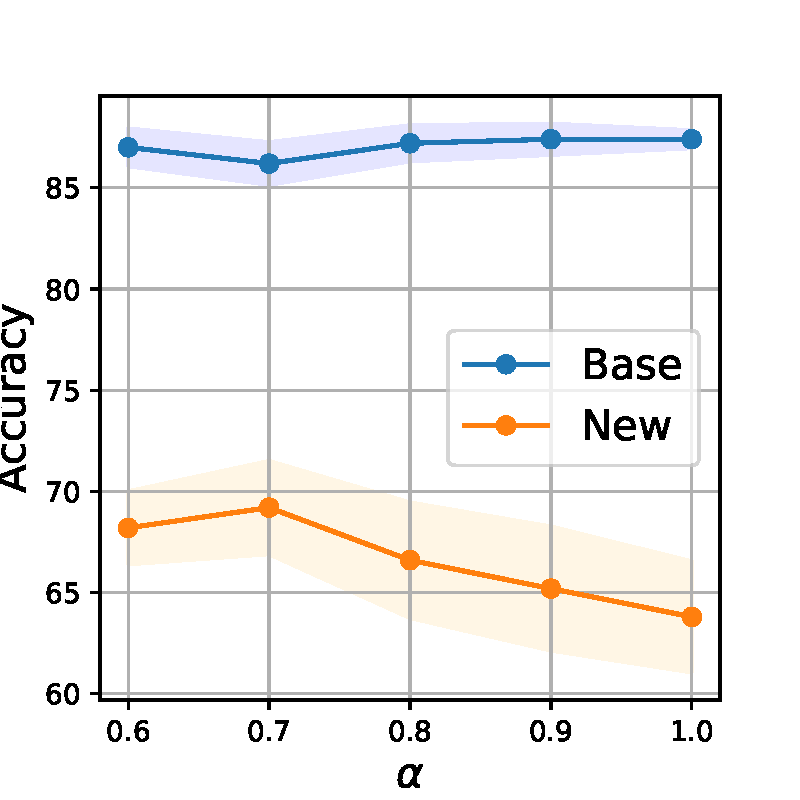}
\caption{UCF101}
\end{subfigure}
\hfill
\begin{subfigure}[h]{0.32\columnwidth}
\includegraphics[width=\linewidth]{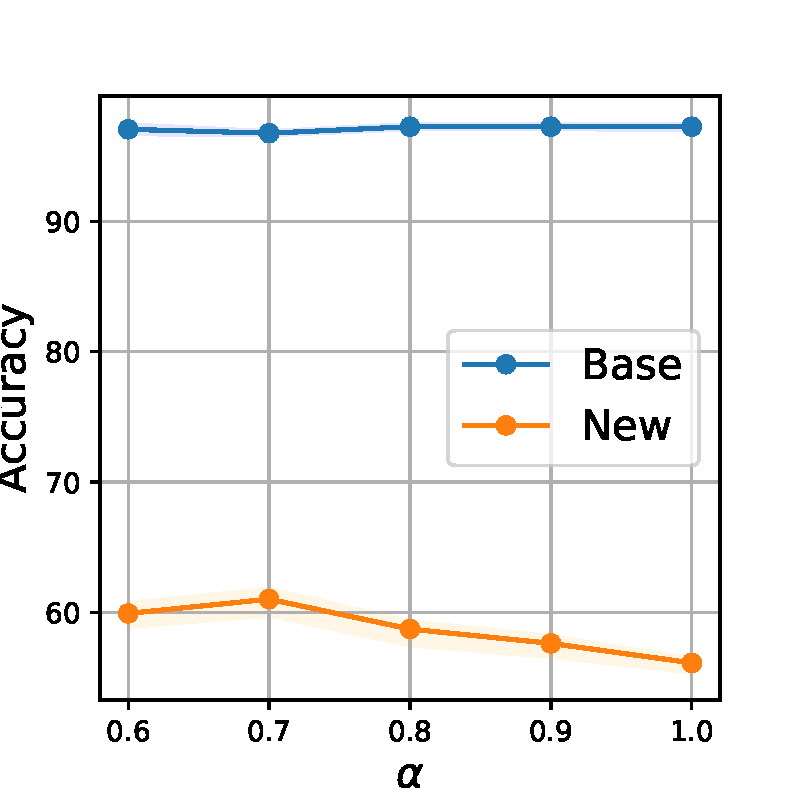}
\caption{Flower102}
\end{subfigure}
\vskip-0.05in
\caption{Sensitivity analysis on $\alpha$, effect of test data-dependent prior. Experiments are replicated over 3 times.}
\label{fig:sense_alpha}
\vskip -0.15in
\end{figure}

\paragraph{Sensitive Analysis of $\alpha$}
We additionally investigate the impact of test data-dependent prior, $p(\theta|x')$, which adapts our posterior distribution to unseen instance. Figure \ref{fig:sense_alpha} shows that adaptation of test data is beneficial for both seen and unseen performances. Additionally, balancing between posterior of seen data and prior of unseen data holds significance in achieving effective generalization for both scenarios.  

\paragraph{Domain Generalization on ImageNet}
Despite the tendency for a potential performance trade-off between the source and target datasets, table \ref{tab:imagenet} demonstrates APP attains the improved performance achieved on both the source and target datasets, highlighting the robustness of APP in dealing with distribution shifts.

\begin{table}[h!]
\caption{Result of domain generalization in ImageNet. Acc represents the accuracy. \textbf{Bold} means the best accuracy.}
\label{tab:imagenet}
\centering
\resizebox{0.85\columnwidth}{!}{
\begin{tabular}{c c|c|c} \toprule
& \textbf{Dataset} & \textbf{Methods} & \textbf{Acc} ($\%$)\\ \midrule
\multirow{4}{*}{Source} &\multirow{4}{*}{ImageNet}  & CoCoOp & $63.13 \pm 0.12 $ \\
&& PLOT  & $63.14 \pm 0.16 $ \\
 & & ProDA &  $62.73 \pm 0.11 $ \\
  & & APP  & $\mathbf{64.50 \pm 0.06}$\\ \midrule
\multirow{16}{*}{Target} &\multirow{4}{*}{ImageNetV2}  & CoCoOp &$55.23 \pm 0.25 $ \\
&& PLOT &  $54.23 \pm 0.45 $  \\
 & & ProDA & $54.97 \pm 0.05 $  \\
  & & APP  &  $\mathbf{57.10 \pm 0.29}$ \\ \cmidrule(lr){2-4}
&\multirow{4}{*}{ImageNet-Sketch}  & CoCoOp & $34.07 \pm 0.46 $ \\
&& PLOT&  $33.93 \pm 0.12 $ \\
 & & ProDA & $34.60 \pm 0.22 $  \\
  & & APP & $\mathbf{35.70 \pm 0.14}$  \\ \cmidrule(lr){2-4}
&\multirow{4}{*}{ImageNet-R}  & CoCoOp & $56.03 \pm 0.38 $ \\
&& PLOT &  $56.86 \pm 0.42 $  \\
 & & ProDA  &$58.57 \pm 0.49$ \\
  & & APP   &$\mathbf{58.70 \pm 0.08}$ \\\cmidrule(lr){2-4}
&\multirow{4}{*}{ImageNet-A}  & CoCoOp & $22.37 \pm 0.09 $ \\
&& PLOT & $22.63 \pm 0.12 $  \\
 & & ProDA  & $23.47 \pm 0.12$ \\
  & & APP  & $\mathbf{23.80 \pm 0.22}$ \\ \bottomrule
\end{tabular}
}
\vskip -0.2in
\end{table}
\section{Conclusion}
We propose the Bayesian framework for prompt learning to consider the uncertainty from few-shot learning scenario, where the image features are possible to be multi-modal and a distribution shift exists between the train and test dataset. We enhance flexibility via Wasserstein Gradient Flow. Furthermore, we propose a novel data-dependent prior distribution that is conditioned on averaged image features. This approach is designed to capture minor modes of image features and facilitate adaptation to previously unseen distributions. We demonstrate substantial performance improvements in various scenarios, including few-shot classifications, domain generalizations, and unseen class generalizations. Additionally, the qualitative analyses indicate that our prompt learning facilitates capturing the multi-modes of image features sparsely. 

\section{Acknowledgments}

This research was supported by AI Technology Development for Commonsense Extraction, Reasoning, and Inference from Heterogeneous Data (IITP) funded by the Ministry of Science and ICT(2022-0-00077).

\bibliography{aaai24}

\newpage
\onecolumn

\section{Supplementary Material}
\subsection{Proof of Proposition 1}

For simplicity, define the $A \coloneqq f(X)$ and $B \coloneqq g(\phi(f(X)),\cdot) $.  Following \cite{NIPS2003_a6ea8471}, we formulate the variational bound for Mutual information as follows:

\begin{equation}
\label{eq:16}
I\left(f(X);g(\phi(f(X)),\cdot)\right) \coloneqq I(A;B) = \E_{p(A,B)} \log {p(B|A) \over p(B) } \ge E_{p(A,B)} \log {q(B|A) \over p(B) }     
\end{equation}
where $q$ is a variational distribution. Following \cite{sordoni2021decomposed}, we define variational distribution $q$ by sampling $B_1,..., B_C$ from the distribution $p(B)$, where $B_i \coloneqq g(\phi(f(X)),y_i)$. 
For formulating the variational distribution, we choose the $B_i$ by the importance weight, which is defined as $ { \exp (sim(A, B_j) / \tau ) \over \sum_{i=1}^C \exp ((sim(A,B_i))/ \tau) } $. 

Therefore, unnormalized variational distribution for $B_1$ is defined as follows: 

\begin{equation}
    \label{eq:17}
    q(B_1|A,B_{2:C})= p(B_1)  { C \cdot \exp (sim(A,B_1) / \tau ) \over \sum_{i=1}^C \exp ((sim(A,B_i))/ \tau) }
\end{equation}

By Jensen's inequality, the lower bound of Eq. \ref{eq:16} is derived as follows: 

\begin{align}
\label{eq:18}
\notag
\E_{p(A,B_1)} \log \E_{p(B_{2:C})} \left({q(B_1|A,B_{2:C}) \over p(B_1) }\right) 
&\ge \E_{p(A,B_1)p(B_{2:C})} \log \left({p(B_1)  { C \cdot \exp (sim(A,B_1) / \tau ) \over \sum_{i=1}^C \exp ((sim(A,B_i))/ \tau) } \over p(B_1) }\right)  \\ 
&=\E_{p(A,B_1)p(B_{2:C})} \log \left({{C \cdot \exp (sim(A,B_1) / \tau ) \over \sum_{i=1}^C \exp ((sim(A,B_i))/ \tau) }  }\right)  \notag \\ 
&=\log C-\mathcal{L}_{CE} (\phi(f(X)),X,Y) 
\end{align}
For training the prior network, we minimize the $\mathcal{L}_{CE}$, which is the upper bound of Mutual information.

\subsection{Derivation from Wasserstein Gradient Flow to Stein Variational Gradient Descent}
Define the functional $F$ as follows:  
\begin{equation}
        F(\mu) \coloneqq D_{KL}(\mu||\pi) 
\end{equation}
, where $\pi \propto \exp(-V(\theta ))$. Then, Wasserstein Gradient is defined as follows:
\begin{equation}
        \nabla {\delta F(\mu) \over \delta \mu }  = \nabla \log {\mu \over \pi}
\end{equation}

Considering Wasserstein Gradient in RKHS with transformation  $\mathcal{K}_{\mu} T(\theta) \coloneqq \E_{\theta' \sim \mu }  [ K({\theta},{\theta'}) T({\theta'})]$, we derive kernelized Wasserstein Gradient as follows: 

\begin{align}
\mathcal{K}_{\mu} \nabla_{\theta'} {\delta F(\mu) \over \delta \mu } &=   \mathcal{K}_{\mu} \nabla_{\theta'} \log {\mu \over \pi} 
\\ &=\int K({\theta},{\theta'}) (- \nabla_{\theta'} \log \pi+ \nabla_{\theta'} \log \mu) \mu(\theta')d\theta' 
\\ &= \int K({\theta},{\theta'}) ( \nabla_{\theta'} V(\theta')+ \nabla_{\theta'} \log \mu) \mu(\theta')d\theta'
\\ &= \int K({\theta},{\theta'}) \nabla_{\theta'} V(\theta') \mu d\theta' -\int \nabla_{\theta'} K({\theta},{\theta'}) \mu d\theta'+ \underbrace{[K({\theta},{\theta'})\mu] \big |_{\|  \theta' \| \rightarrow \infty}}_{=0}
\end{align}

By the Continuity equation, evolving path of $\theta$ is derived as follows:

\begin{align}
    \notag
    \partial_t \theta_t&=- \mathcal{K}_{\mu} \nabla_{\theta'} ({\delta F(\mu) \over \delta \mu} ) \\ & =- [\int K({\theta},{\theta'}) \nabla_{\theta'} V(\theta') \mu d\theta' -\int \nabla_{\theta'} K({\theta},{\theta'}) \mu d\theta']
\end{align}

By discretization, we formulate the following update rule: 

\begin{equation}
    \theta_{t+1}^i=\theta_{t}^i - {h \over M} \sum_{j=1}^M  [K(\theta_t^i,\theta_t^j) \nabla_{\theta_t^j} V(\theta_t^j)-\nabla_{\theta_t^j}K(\theta_t^i,\theta_t^j)]
\end{equation}

\subsection{Implementation Details}

We adopted the training settings from the work of \cite{chen2023plot}. In our experiments, we used a batch size of 32 for the Oxford-Flowers, FGVC-Aircraft, and Stanford-Cars datasets, while the batch size for other datasets was fixed at 128. The prior network was trained for 10, 20, 20, 40, and 40 shots for each respective dataset. We employed the SGD optimizer with an initial learning rate of 0.002, which was annealed using the CosineAnnealing schedule. All experiments were conducted on a single NVIDIA A100 GPU core.

For evaluation, we considered the model output within a Bayesian framework, taking into account the posterior distribution $p(\theta|X, Y)$. Thus, predictive distribution for test data point $x'$ is defined as follows:

\begin{equation}
    p(y'|x',\mathcal{D})=\int p(y'|x',\theta)p(\theta|\mathcal{D},x') d\theta 
\end{equation}
, where $\mathcal{D} \coloneqq \{X,Y\}$

Table \ref{tab:test_acc} indicates a performance comparison between ours and other multi-prompts methods.

\begin{table*}[h!]
\caption{Result of Few-shot Classification. We report the mean accuracy by conducting three-replicated experiments. }

\label{tab:test_acc}
\centering
\resizebox{0.8\textwidth}{!}{
\begin{tabular}{c|c|c|c|c|c|c}
\hline 
 Dataset  & Methods         & 1 shot & 2 shots & 4 shots & 8 shots & 16 shots   \\ 
\hline
\multirow{3}{*}{Caltech101} & PLOT \cite{chen2023plot}         & $89.72 \pm 0.44 $  & $90.48 \pm 0.22 $  & $90.99 \pm 0.30 $  & $91.23 \pm 0.46 $ & $92.29 \pm 0.22 $   \\
 & ProDA \cite{lu2022prompt}          & $89.14 \pm 0.19 $  & $90.08 \pm 0.50 $  & $90.94 \pm 0.19 $  & $91.83 \pm 0.27 $ & $92.47 \pm 0.08 $   \\
  & Ours       & $\mathbf{90.30 \pm 0.33} $  & $ \mathbf{90.87 \pm 0.25}$  & $\mathbf{91.05 \pm 0.45}$  & $\mathbf{91.92 \pm 0.08} $ & $\mathbf{92.89 \pm 0.22}$   \\
\hline

\multirow{3}{*}{DTD} & PLOT \cite{chen2023plot}      & $46.94 \pm 1.89 $  & $51.46 \pm 2.27$  & $55.95 \pm 0.74 $  & $61.68 \pm 0.34 $ & $65.27 \pm 0.23 $   \\
 & ProDA \cite{lu2022prompt}       & $47.52 \pm 1.69 $  & $\mathbf{52.17 \pm 1.89} $  & $55.85 \pm 0.92 $  & $62.74 \pm 0.38 $ & $66.27 \pm 1.11$   \\
  & Ours       & $\mathbf{49.09 \pm 2.29} $  & $51.73\pm 3.47$  & $\mathbf{58.06 \pm 0.69} $  & $\mathbf{63.42 \pm 0.12} $ & $\mathbf{66.57 \pm 0.07}$   \\
\hline

\multirow{3}{*}{EuroSAT} & PLOT \cite{chen2023plot}      & $54.15 \pm 2.04 $  & $63.64 \pm 3.08$  & $74.91 \pm 2.13 $  & $78.55 \pm 1.33 $ & $84.06 \pm 0.91 $   \\
 & ProDA \cite{lu2022prompt}       & $49.89 \pm 2.37 $  & $62.37 \pm 1.41 $  & $70.38 \pm 1.73 $  & $77.20 \pm 1.97 $ & $80.62 \pm 1.51 $   \\
  & Ours       & $\mathbf{60.04 \pm 2.08} $  & $\mathbf{66.02 \pm 2.70} $  & $\mathbf{74.08 \pm 2.23} $  & $\mathbf{77.87 \pm 2.51} $ & $\mathbf{84.08 \pm 1.1} $   \\
\hline

\multirow{3}{*}{FGVC-Aircraft}  & PLOT \cite{chen2023plot}  & $18.22 \pm 0.59 $  & $19.26 \pm 0.96 $  & $22.39 \pm 1.41 $  & $27.58 \pm 0.83 $ & $32.28 \pm 0.43 $   \\
 & ProDA \cite{lu2022prompt}         & $\mathbf{20.25 \pm 0.88 }$  & $\mathbf{22.45 \pm 0.37 }$  & $24.76 \pm 1.00 $  & $28.46 \pm 0.43 $ & $32.84 \pm 0.34 $   \\
  & Ours       & $18.10\pm 0.92$  & $21.87 \pm 1.21$  & $\mathbf{25.65 \pm 0.59} $  & $\mathbf{29.40  \pm 0.42} $ & $\mathbf{33.80 \pm 0.53}$   \\
\hline

\multirow{3}{*}{Food101}  & PLOT \cite{chen2023plot}  & $77.87 \pm 0.13 $  & $77.78 \pm 0.37 $  & $77.20 \pm 0.37 $  & $75.42 \pm 0.09 $ & $77.18 \pm 0.16 $   \\
 & ProDA \cite{lu2022prompt}         & $78.65 \pm 0.17 $  & $79.04 \pm 0.17 $  & $\mathbf{79.47 \pm 0.17} $  & $78.88 \pm 0.22 $ & $79.82 \pm 0.10$   \\
  & Ours       & $\mathbf{78.87 \pm 0.17}$  & $\mathbf{79.25 \pm 0.06 } $  & $79.43 \pm 0.10 $  & $\mathbf{79.49  \pm 0.17} $ & $\mathbf{79.83 \pm 0.06}$   \\
\hline

\multirow{3}{*}{ImageNet}  & PLOT \cite{chen2023plot}  & $59.32 \pm 0.58 $  & $59.81 \pm 0.25 $  & $61.12 \pm 0.14 $  & $61.91 \pm 0.09 $ & $63.14 \pm 0.16 $   \\
 & ProDA \cite{lu2022prompt}         & $61.65 \pm 0.46 $  & $62.20 \pm 0.11 $  & $61.96 \pm 0.30 $  & $61.78 \pm 0.29 $ & $62.73 \pm 0.11 $   \\
  & Ours       & $\mathbf{62.11\pm 0.26}$  & $\mathbf{62.50 \pm 0.10 } $  & $\mathbf{63.05 \pm 0.15} $  & $\mathbf{63.70  \pm 0.17} $ & $\mathbf{64.50 \pm 0.06}$   \\
\hline

\multirow{3}{*}{Oxford Flowers}  & PLOT \cite{chen2023plot}  & $72.32 \pm 0.94 $  & $82.65 \pm 0.38 $  & $88.13 \pm 0.27 $  & $92.56 \pm 0.02 $ & $95.64 \pm 0.28 $   \\
 & ProDA \cite{lu2022prompt}         & $70.56 \pm 0.48 $  & $83.73 \pm 0.50 $  & $88.94 \pm 0.30 $  & $93.30 \pm 0.35 $ & $95.47 \pm 0.18 $   \\
  & Ours       & $\mathbf{74.42\pm 0.58}$  & $\mathbf{84.23 \pm 0.14 } $  & $\mathbf{89.13 \pm 0.31} $  & $\mathbf{93.38  \pm 0.09} $ & $\mathbf{95.67 \pm 0.09}$   \\
\hline

\multirow{3}{*}{Oxford Pets}  & PLOT \cite{chen2023plot}  & $87.35 \pm 0.67 $  & $87.16 \pm 0.37 $  & $88.25 \pm 0.58 $  & $87.39 \pm 0.39 $ & $87.20 \pm 0.20 $   \\
 & ProDA \cite{lu2022prompt}         & $88.76 \pm 0.08 $  & $88.15 \pm 0.39 $  & $89.17 \pm 0.41 $  & $\mathbf{89.97 \pm 0.23 }$ & $89.61 \pm 0.36$   \\
  & Ours       & $\mathbf{88.97\pm 0.52}$  & $\mathbf{88.28 \pm 0.39 } $  & $\mathbf{89.48 \pm 0.28} $  & $89.79  \pm 0.16 $ & $\mathbf{89.73 \pm 0.63}$   \\
\hline

\multirow{3}{*}{Stanford Cars}  & PLOT \cite{chen2023plot}  & $56.21 \pm 0.78 $  & $57.35 \pm 0.27 $  & $63.35 \pm 0.54 $  & $67.60 \pm 0.05 $ & $73.78 \pm 0.11 $   \\
 & ProDA \cite{lu2022prompt}         & $59.55 \pm 0.13 $  & $62.08 \pm 0.65 $  & $66.32 \pm 0.18 $  & $71.11 \pm 0.21 $ & $74.86 \pm 0.21 $   \\
  & Ours       & $\mathbf{59.65 \pm 0.41}$  & $\mathbf{63.39 \pm 0.35 } $  & $\mathbf{67.18 \pm 0.30} $  & $\mathbf{71.86  \pm 0.11} $ & $\mathbf{76.14 \pm 0.23}$   \\
\hline

\multirow{3}{*}{Sun397}  & PLOT \cite{chen2023plot}  & $63.03 \pm 0.32 $  & $62.10 \pm 0.50 $  & $65.60 \pm 0.44 $  & $66.90 \pm 0.25 $ & $69.76 \pm 0.29 $   \\
 & ProDA \cite{lu2022prompt}         & $\mathbf{63.99 \pm 0.15 }$  & $65.56 \pm 0.40$  & $\mathbf{67.79 \pm 0.49}$  & $\mathbf{69.50 \pm 0.13}$ & $\mathbf{71.61 \pm 0.18}$   \\
  & Ours       & $\mathbf{63.99\pm 0.20}$  & $\mathbf{65.59 \pm 0.19}$  & $67.01 \pm 0.36$  & $69.46  \pm 0.08$ & $71.59 \pm 0.19$   \\
\hline

\multirow{3}{*}{UCF101}  & PLOT \cite{chen2023plot}  & $64.31 \pm 0.36 $  & $67.64 \pm 0.42 $  & $70.87 \pm 0.32 $  & $75.77 \pm 0.25 $ & $77.58 \pm 0.29 $   \\
 & ProDA \cite{lu2022prompt}         & $64.44 \pm 0.68 $  & $67.80 \pm 0.11 $  & $70.78 \pm 0.23 $  & $\mathbf{77.01 \pm 0.35} $ & $\mathbf{79.34 \pm 0.68}$   \\
  & Ours       & $\mathbf{65.83 \pm 0.84}$  & $\mathbf{69.05 \pm 0.19} $  & $\mathbf{71.79 \pm 0.13} $  & $76.71
  \pm 0.60 $ & $79.07 \pm 0.12$   \\
\specialrule{.2em}{.1em}{.1em}

\multirow{3}{*}{Average}  & PLOT \cite{chen2023plot}  & $62.68 \pm 0.09 $  & $65.39 \pm 0.48 $  & $68.98 \pm 0.24 $  & $71.51 \pm 0.10 $ & $74.38 \pm 0.06 $   \\
 & ProDA \cite{lu2022prompt}         & $63.13 \pm 0.26 $  & $66.88 \pm 0.22 $  & $69.67 \pm 0.16 $  & $72.89 \pm 0.25 $ & $75.06 \pm 0.05$   \\
  & Ours       & $\mathbf{64.67 \pm 0.26}$  & $\mathbf{67.53 \pm 0.50 } $  & $\mathbf{70.54 \pm 0.15} $  & $\mathbf{73.36  \pm 0.24}$ & $\mathbf{75.81 \pm 0.06}$   \\
\hline

\end{tabular}
}
\end{table*}
\subsection{Semantics of Prompts}

To analyze the semantic meaning of the learned context vectors in our approach, we performed a study where we extracted the nearest words in the embedding space of each context vector. While not all vectors have a straightforward semantic interpretation in continuous space, we observed that certain components of each context vector exhibited clear semantic relevance and effectively described the image data.

Table \ref{tab:semantics} demonstrates that each context vector is trained in a direction that corresponds to specific aspects of the image data. This finding indicates that our model has successfully learned to capture meaningful representations of the image content within the context vectors. For instance, context 1 contains words such as "americanair" and "usnavy," which are associated with civil aircraft and fighter aircraft, respectively, in the dataset.
\begin{table}[h!]
\caption{Interpretable and closest words to our context vectors for FGVC-Aircraft. We report one of the top 4 words, which are interpretable. - means that there are no interpretable words and bold means highly relevant words to datasets. }
\label{tab:semantics}
\centering
\resizebox{0.6\textwidth}{!}{
\begin{tabular}{c|c|c|c|c}
\hline 
 Number& Context 1&Context 2&Context 3&Context 4\\ 
\hline
1 &sculpt&calling &salazar & -\\
2 &espn&byte & grassroots & bucketlist\\
3 &postponed& turquoise& installation & below\\
4 &grocery& \textbf{wig}& \textbf{followed}&accepted \\
5 &taking& shells&translates &turkish \\
6 &\textbf{usnavy}&likes&administrator &fair \\
7 &fort& -& centred& \textbf{qantas}\\
8 &staten& \textbf{fo}&blaze &modern \\
9 &\textbf{rusher}& tue& attempting&serve\\
10 &wielding&ata&- &\textbf{crossing} \\
11 &trick&brown &blowing &- \\
12 &pate&vfl&donkey & occasion\\
13 &\textbf{americanair}&- & legendary&times \\
14 &eight&ima&reunite & inevit\\
15 &..?&thanku&\textbf{flew} &letting \\
16 &brox &\textbf{facing} &!&grand \\
\hline 

\end{tabular}
}
\vskip -0.2in
\end{table}

\subsection{Empirical comparison with BPL}

Table \ref{tab:bpl} (unseen classes generalization experiments) indicates that APP is more efficient in time and memory complexity with more robust performance, harmonic mean (accuracies between base and new classes). This gain demonstrates that the Gaussian variational distribution of BPL is not sufficient to capture the whole multimodes of image features. In addition, the conditional posterior assumption is not scalable, when the batch size is set to high.

\begin{table}[h!] 
    \centering
    \resizebox{0.6\textwidth}{!}{  
    \begin{tabular}{c|c|c|c|c|c} \toprule
        \textbf{Dataset}&\textbf{Method} & \textbf{BS}& \textbf{Memory (GB)}& \textbf{Time (hr)} & \textbf{H (\%)}  \\ \midrule
        \multirow{4}{*}{Caltech101} &\multirow{2}{*}{BPL} &1 &11.0 & 5.75 & $92.3$ \\
          & &128&-&-&- \\
        
        &\multirow{2}{*}{APP (Ours)} &1 &$\mathbf{10.6}$& $\mathbf{2.50}$ & $\mathbf{93.0} $ \\
          & & 128&$\mathbf{10.8}$&$\mathbf{0.25}$&$\mathbf{93.0}$ \\
        \bottomrule

        \multirow{4}{*}{Oxford pets} &\multirow{2}{*}{BPL} &1 &7.4 &1.5 & $91.5$ \\
          & &128&-&-&- \\
        
        &\multirow{2}{*}{APP (Ours)} &1 &$\mathbf{7.0}$& $\mathbf{0.83}$ & $\mathbf{91.7} $ \\
          & & 128&$\mathbf{8.2}$&$\mathbf{0.06}$&$\mathbf{92.4}$ \\
        \bottomrule

        \multirow{4}{*}{UCF101} &\multirow{2}{*}{BPL} &1 &11.5 & 9.26 & $72.8$ \\
          & &128&-&-&- \\
        
        &\multirow{2}{*}{APP (Ours)} &1 &$\mathbf{7.0}$& $\mathbf{3.25}$ & $\mathbf{73.3} $ \\
          & & 128&$\mathbf{8.5}$&$\mathbf{0.1}$&$\mathbf{76.8}$ \\
        \bottomrule
        
    \end{tabular} 
    }
    \caption{Results of BPL in unseen classes generalization experiments. We report the performance of three replicated experiments. The symbol "-" indicates the training failure due to the memory issue.}
    \label{tab:bpl}
    \end{table}

\end{document}